%% file: my_final_arxiv.tex
\pdfoutput=1

\documentclass[11pt]{article}

\usepackage[preprint]{acl}

\usepackage{times}
\usepackage{latexsym}
\usepackage{algorithm}
\usepackage{algpseudocode}
\usepackage{algorithmicx}   %

\usepackage[T1]{fontenc}

\usepackage[utf8]{inputenc}

\usepackage{microtype}

\usepackage{inconsolata}

\usepackage{graphicx}
\usepackage{amsmath}
\usepackage{amssymb}
\usepackage{mathtools}
\usepackage{amsthm}
\usepackage{amssymb}
\usepackage{hyperref}
\usepackage{url}
\usepackage{multirow}
\usepackage{multicol}
\usepackage{xcolor}
\usepackage{colortbl}
\usepackage{booktabs}
\usepackage{subfigure}
\usepackage{siunitx}
\usepackage{subcaption}
\usepackage{tablefootnote}
\usepackage{enumitem}

\definecolor{BLUE}{HTML}{395A9D}
\definecolor{ORANGE}{HTML}{F36847}

\title{Measuring Data Diversity for Instruction Tuning: \\A Systematic Analysis and A Reliable Metric}

\author{
    \bf{\normalsize
    Yuming Yang$^{1}$\thanks{Equal Contribution.},\ \
    Yang Nan$^{1}$\footnotemark[1],\ \
    Junjie Ye$^{1}$,\ \
    Shihan Dou$^{1}$,}\\
    \bf{\normalsize
    Xiao Wang$^{1}$,\ \
    Shuo Li$^{1}$,\ \
    Huijie Lv$^{1}$,\ \
    Mingqi Wu$^{1}$,}\\
    \bf{\normalsize
    Tao Gui$^{2,3,4}$\thanks{Corresponding Authors.},\ \
    Qi Zhang$^{1,3,4}$\footnotemark[2],\ \
    Xuanjing Huang$^{1,3,4}$\footnotemark[2]}\\
  {$^1$ \normalsize College of Computer Science and Artificial Intelligence, Fudan University}\\
  {$^2$ \normalsize Institute of Modern Languages and Linguistics, Fudan University}\\
  {$^3$ \normalsize Institute of Trustworthy Embodied Artificial Intelligence, Fudan University}\\
  {$^4$ \normalsize Shanghai Collaborative Innovation Center of Intelligent Visual Computing}\\
  \texttt{\normalsize yumingyang23@m.fudan.edu.cn} \ \ \ \texttt{\normalsize \{qz,tgui,xjhuang\}@fudan.edu.cn} \\
}

\begin{document}
\maketitle
\begin{abstract}
\input{latex/sections/abstract}

\end{abstract}

\input{latex/sections/intro}

\input{latex/sections/existing}

\input{latex/sections/novelsum}

\input{latex/sections/simulation}

\input{latex/sections/experiment}

\input{latex/sections/novelselect}
\input{latex/sections/discussion}
\input{latex/sections/related}

\section{Conclusion}
In this paper, we investigate the fundamental problem of precisely measuring dataset diversity for instruction tuning and propose \textit{NovelSum}, a reliable diversity metric that correlates well with model performance. Inspired by our systematic analysis of existing diversity metrics, \textit{NovelSum} jointly considers inter-sample distances and information density to effectively capture dataset diversity, achieving superior correlations with model performance compared to previous metrics. Based on \textit{NovelSum}, We further develop a data selection strategy, \textit{NovelSelect}, whose remarkable performance validates the practical significance of \textit{NovelSum}.

\section*{Limitations}
\begin{itemize}[itemsep=3pt, parsep=0pt, topsep=3pt]
    \item Although our study systematically analyzes both existing and proposed metrics through extensive fine-tuning experiments, we focus on Qwen-2.5-7B and LLaMA-3-8B as the backbone LLMs, excluding larger models and other series due to resource constraints.
    \item While we strive to employ comprehensive benchmarks to evaluate instruction tuning performance, the currently available test data may still fall short of fully capturing the diversity of real-world use cases. As a result, the beneficial effects of data diversity on model capabilities may be underrepresented in benchmark results.
    \item As previously noted, diversity measurements on downstream IT tasks may differ from our analysis in the general IT setting, suggesting the need for further study.
\end{itemize}

\section*{Acknowledgements}
The authors wish to thank the anonymous reviewers for their helpful comments. This work was partially funded by National Natural Science Foundation of China (No.62476061, 62206057, 61976056), Shanghai Rising-Star Program (23QA1400200), and Natural Science Foundation of Shanghai (23ZR1403500).

\bibliography{custom}

\clearpage
\appendix
\input{latex/sections/appendix}

\end{document}

%% file: latex/sections/abstract.tex
Data diversity is crucial for the instruction tuning of large language models. 
Existing studies have explored various diversity-aware data selection methods to construct high-quality datasets and enhance model performance. 
However, the fundamental problem of precisely defining and measuring data diversity remains underexplored, limiting clear guidance for data engineering. 
To address this, we systematically analyze 11 existing diversity measurement methods by evaluating their correlation with model performance through extensive fine-tuning experiments. 
Our results indicate that a reliable diversity measure should properly account for both inter-sample differences and the information density in the sample space.
Building on this, we propose \textit{NovelSum}, a new diversity metric based on sample-level "novelty." 
Experiments on both simulated and real-world data show that \textit{NovelSum} accurately captures diversity variations and achieves a 0.97 correlation with instruction-tuned model performance, highlighting its value in guiding data engineering practices. 
With \textit{NovelSum} as an optimization objective, we further develop a greedy, diversity-oriented data selection strategy that outperforms existing approaches, validating both the effectiveness and practical significance of our metric.
The code is available at \url{https://github.com/UmeanNever/NovelSum}.

%% file: latex/sections/intro.tex
\section{Introduction}
Instruction tuning (IT) fine-tunes pretrained large language models (LLMs) with annotated instruction data, enabling them to follow human instructions and perform various tasks effectively \cite{sanh2022multitask, zhang2023instruction}. Recent studies indicate that small-scale, high-quality datasets can outperform larger ones in IT performance \cite{chen2023maybe-Kcentergreedy, zhou2024lima}, with data diversity playing a crucial role in achieving optimal results \cite{liu2023makes, bukharin2023data-QDIT, zhang2024instruction, yang2024beyond}. Consequently, various diversity-aware data selection methods have emerged \cite{qin2024unleashing, wang2024survey}, driven by different interpretations of data diversity.

\begin{figure}[t!]
    \centering
        \includegraphics[width=\linewidth]{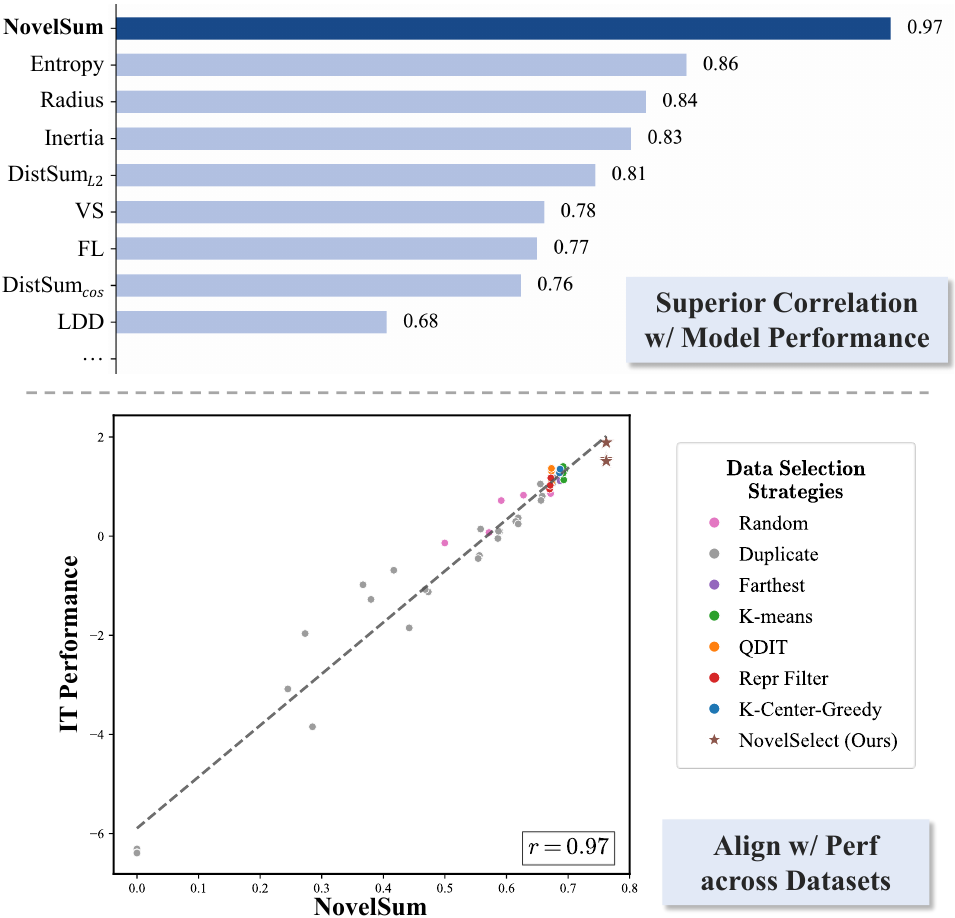}
    \caption{Our diversity metric, \textit{NovelSum}, exhibits superior correlation with model performance compared to existing metrics across IT datasets constructed with various data selection strategies.}
    \label{fig:head}
    \vspace{-6mm}
\end{figure}

However, the fundamental problem of precisely defining and measuring data diversity remains underexplored. This ambiguity has turned data engineering for diversity into a black-box process, leading to data selection methods that often fail to generalize and, at times, perform worse than random selection \cite{xia2024rethinking,diddee2024chasing}. 
While some diversity metrics have been introduced in IT research \cite{bukharin2023data-QDIT,wang2024diversity-logD}, a comprehensive evaluation and comparative analysis are still needed to identify a reliable metric that strongly correlates with fine-tuning performance in practice. 

To this end, we systematically analyze and evaluate the reliability of 11 existing diversity metrics through extensive experiments. 
Using various mainstream diversity-oriented data selection methods, we construct 53 IT datasets and fine-tune models accordingly. We then measure dataset diversity using existing metrics and assess their correlation with model performance. By analyzing the limited correlation of existing metrics, we find that: (1) \textbf{A reliable diversity metric must capture differences between samples} to reflect each sample's information uniqueness. Moreover, differences between neighboring samples are more critical for overall diversity but can be overshadowed by variations in distant samples. (2) \textbf{Measuring differences between samples should account for both semantic similarity and the uneven distribution of information in space.} In high-density domains like math and code, semantically similar samples can still contain substantial unique information and should therefore be considered more diverse.

Building on these insights, we propose \textit{NovelSum}, a diversity metric that jointly considers inter-sample differences and uneven information density. Specifically, we define dataset diversity as the sum of each sample's unique contribution to overall information, termed "novelty". Just as a research paper’s novelty is judged by its distinction from related work based on field-specific standards, we compute a sample's novelty as the proximity-weighted sum of its differences from other samples in the dataset. These differences are measured using density-aware distances, which capture both semantics and local information density.

To validate the effectiveness of \textit{NovelSum}, we conduct both a visualized simulation study and real-world correlation experiments using two different LLMs. 
The results show that \textit{NovelSum} accurately captures diversity variations and strongly correlates with instruction-tuned model performance, achieving Pearson's $r=0.98$ and Spearman's $r=0.95$, outperforming other metrics.
This demonstrates \textit{NovelSum}'s potential to effectively guide data engineering practices.
Furthermore, we develop \textit{NovelSelect}, a greedy, diversity-oriented data selection strategy that uses \textit{NovelSum} as the optimization objective. Experimental results confirm its superior performance compared to other approaches.

Our main contributions are three-fold:
\begin{itemize}[itemsep=3pt, parsep=0pt, topsep=3pt]
\item We systematically analyze and evaluate the reliability of existing diversity metrics for instruction tuning by computing their correlation with model performance, thereby unveiling pathways to a more reliable metric.
\item We propose \textit{NovelSum}, a diversity metric that captures both inter-sample differences and information density, achieving a strong correlation with instruction-tuning performance, substantially exceeding previous metrics.
\item We develop \textit{NovelSelect}, a diversity-oriented data selection strategy based on \textit{NovelSum}, which outperforms existing methods and further validates \textit{NovelSum}'s effectiveness and practical value in instruction tuning.

\end{itemize}

%% file: latex/sections/existing.tex
\section{Evaluating Existing Diveristy Metrics}
\label{sec:existing}
We begin by evaluating the correlation between existing diversity metrics and instruction-tuned model performance, identifying limitations to inform the design of a more reliable metric.

Our evaluation follows four steps: 
(1) Construct multiple IT datasets, each denoted as $\mathcal{X}^{(s)}$, using different data selection strategies from the full data source $\mathcal{X}^{all}$. 
(2) Measure dataset diversity using existing metrics, denoted as $\mathcal{M}_{t}(\mathcal{X}^{(s)})$. 
(3) Fine-tune LLMs on each dataset and evaluate their performance, $\mathcal{P}^{(s)}$, using IT benchmarks.
(4) Analyze the correlation between each diversity metric and model performance, denoted as $r_{\mathcal{M}_{t},\ \mathcal{P}}$.

\subsection{Existing Diversity Metrics}
We use 11 existing diveristy metrics for the analysis, categoried into three main types:

\paragraph{Lexical Diversity}  
A classical way to measure textual diversity is by analyzing vocabulary usage, where a higher proportion of unique words indicates greater diversity. Two widely used metrics are the \textbf{Type-Token Ratio} (TTR) \cite{richards1987type-TTR} and \textbf{vocd-D} \cite{malvern2004lexical-vocd}, with details in the Appendix \ref{app:exist}.

\paragraph{Distance-based Semantic Diversity}  
Recent studies primarily measure dataset diversity based on the semantics of individual samples, often represented as embeddings $emb(\cdot)$ from language models like BERT. A common approach quantifies diversity by computing distances between samples using their embeddings, encouraging heterogeneity. 
For example, a straightforward metric sums the pairwise distances among all samples in a dataset:
\begin{equation}
\label{eq:distsum}
    \mathcal{M}_{DistSum}(\mathcal{X}) = \sum_{x_i, x_j \in \mathcal{X}, i \neq j} \Delta(x_i, x_j),
\end{equation}  
where $\Delta(\cdot, \cdot)$ denotes the distances between two samples. Specifically, \textbf{DistSum$_{cosine}$} uses cosine distance and \textbf{DistSum$_{L2}$} uses Euclidean distance.
Beyond simple summation, more refined metrics are proposed. The \textbf{KNN distance} \cite{stasaski2020more-KNN, stasaski2022semantic-KNN} measures the average distance of each sample to its $k$-nearest neighbor, ensuring sample uniqueness: 
\begin{equation}  
    \mathcal{M}_{KNN}(\mathcal{X}) = \frac{1}{|\mathcal{X}|} \sum_{i=1}^{|\mathcal{X}|} \Delta(x_i, N_k(x_i)),
\end{equation}
where $N_k(x_i)$ denotes the k-th closest neighbor of $x_i$, typically with $k=1$. 
We also compute \textbf{Cluster Inertia} \cite{du2019boosting-Inertia}, \textbf{Vendi Score} \cite{pasarkar2023cousins-Vendi}, \textbf{Radius} \cite{lai2020diversity-Radius} and \textbf{Log Determinant Distance} (LDD) \cite{wang2024diversity-logD}; see Appendix~\ref{app:exist} for details.

\paragraph{Distribution-based Semantic Diversity}  
Another notable class of metrics measures diversity from a distributional perspective, assessing how well a selected dataset $\mathcal{X}$ represents the overall sample (semantic) space of $\mathcal{X}^{all}$. 
One example is the \textbf{Facility Location} (FL) function \cite{farahani2009facility-FL}, which defines a dataset as diverse if each sample in $\mathcal{X}^{all}$ has a close representative in $\mathcal{X}$, ensuring thorough coverage of space:
\begin{equation}
    \mathcal{M}_{FL}(\mathcal{X}) = \sum_{x_j \in \mathcal{X}^{all}}\min_{x_i \in \mathcal{X}} \Delta(x_i, x_j)
\end{equation}  
Another feasible metric, \textbf{Partition Entropy} \cite{han2022measuring}, captures how evenly the selected dataset spans the sample space. It applies K-means to partition $\mathcal{X}^{all}$ into $K$ clusters and calculates the entropy of the cluster distribution for $\mathcal{X}$.
\begin{equation}
\setlength\abovedisplayskip{8pt}%
\setlength\belowdisplayskip{8pt}
    \mathcal{M}_{Entropy}(\mathcal{X}) = -\sum_{k=1}^{K} p_k \log p_k,  
\end{equation}
where $p_k$ is the proportion of selected samples in cluster $k$. Higher entropy indicates greater distributional uncertainty and a more balanced dataset.

\subsection{IT Dataset Construction and Benchmark}

Focusing on general IT, we follow \citealp{liu2023makes} to construct our IT data source by combining WizardLM \cite{xu2024wizardlm}, ShareGPT \cite{chiang2023vicuna-ShareGPT}, and UltraChat \cite{ding2023enhancing-UltraChat}, denoted as $\mathcal{X}^{all}$. We extract embeddings for each sample in $\mathcal{X}^{all}$. 
See Appendix~\ref{app:preprocess} for preprocessing details.

We then apply several representative diversity-aware data selection strategies to curate IT datasets, yielding subsets $\mathcal{X}^{(s)} \subset \mathcal{X}^{all}$. 
To minimize the influence of factors beyond diversity, we control for sample quality differences across datasets by removing anomalous source samples and excluding any data quality filters during selection. We also fix the dataset size at 10,000 samples. 
The strategies used are: \textbf{K-Center-Greedy} \cite{sener2017active-Kcentergreedy, du2023mods-Kcentergreedy, wu2023self-Kcentergreedy}, which iteratively selects the sample farthest from the current coreset; \textbf{Repr Filter} \cite{liu2023makes}, which improves $\mathcal{M}_{KNN}$ by applying a minimum distance threshold when adding samples into the coreset; \textbf{QDIT} \cite{bukharin2023data-QDIT}, which optimizes diversity by serially selecting the data point that maximizes $\mathcal{M}_{FL}$; \textbf{K-means} \cite{song2024iterselecttune-Kmeans, ge2024clustering}, which partitions samples into clusters and evenly select samples from each; and baselines, including \textbf{Random} selection and \textbf{Farthest}, which ranks samples by their total distances to others and selects the most distant ones. Additionally, we construct datasets with varying amounts of \textbf{Duplicate} samples to simulate low-diversity datasets. Each strategy is run at least three times to ensure robustness, yielding 53 IT datasets. 
Details on dataset construction are provided in Appendix \ref{app:ds}. 

We fine-tune LLaMA-3-8B \cite{dubey2024llama} on these datasets and evaluate model performance using two popular IT benchmarks: MT-bench \cite{zheng2023judging-MT-Bench} and AlpacaEval \cite{li2023alpacaeval}. 
See Appendix~\ref{app:eval} for details and rationale of the benchmarks.
To jointly consider both benchmarks, we normalize the results into Z-scores and compute the aggregated performance as 
\begin{equation}
\label{eq:perf}
    \mathcal{P}^{(s)} = z^{(s)}_{MT-bench} + z^{(s)}_{AlpacaEval}
\end{equation}

\begin{figure*}[t!]
    \centering
        \includegraphics[width=\linewidth]{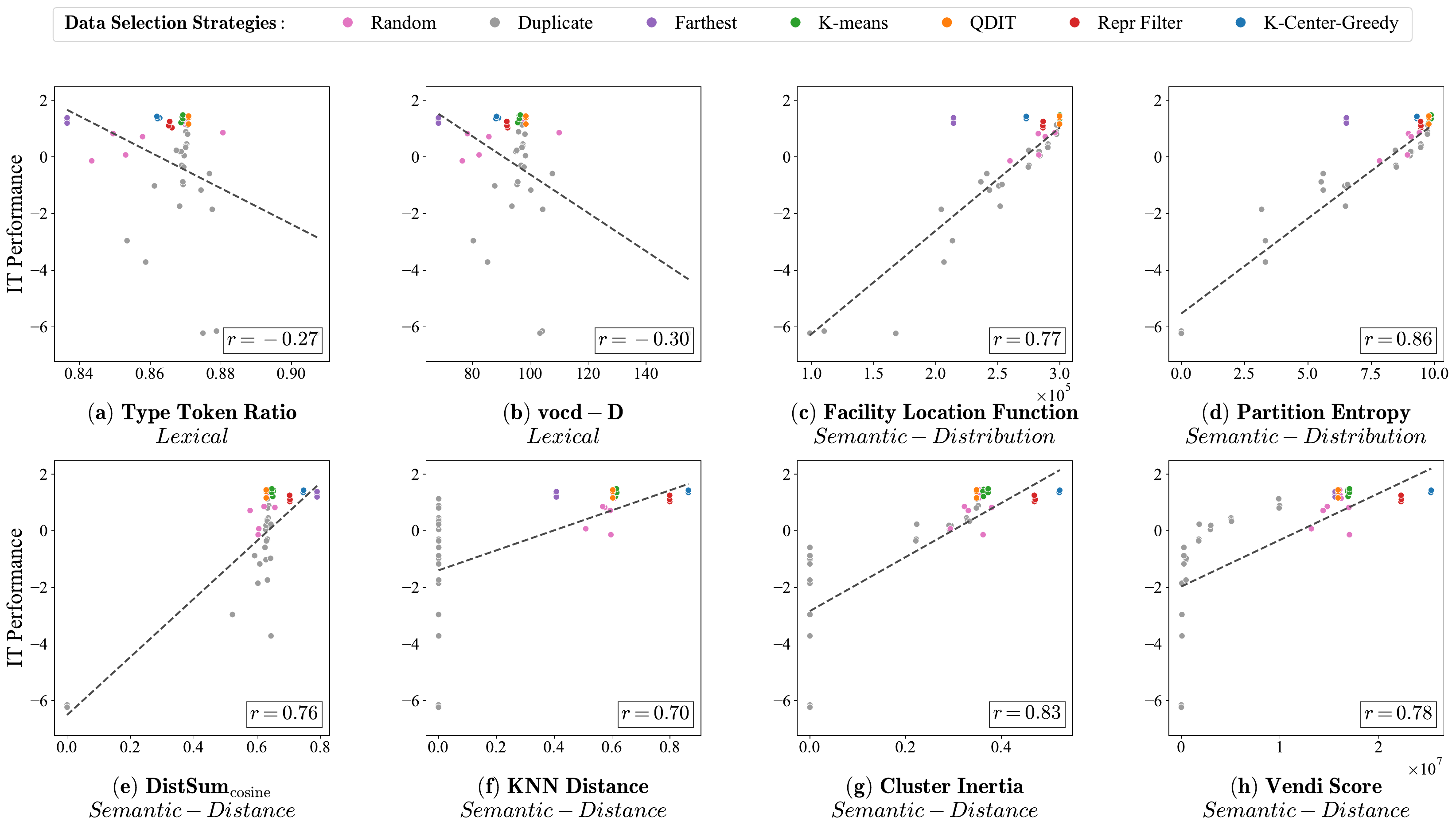}
    \caption{Evaluating existing diversity metrics based on their correlation (Eq. \ref{eq:cor}) with IT performance (Eq. \ref{eq:perf}). The X-axis represents diversity measurements. Each point corresponds to a 10k IT dataset constructed using different strategies. Abnormal points highlight the limitations of current metrics and inspire the development of new ones.}
    \label{fig:exist}
    \vspace{-4.5mm}
\end{figure*}

\subsection{Correlation Analysis}
\label{subsec:analysis}
Finally, we compute the correlation between each diversity metric $\mathcal{M}_{t}$ and model performance $\mathcal{P}$ by averaging their Pearson and Spearman coefficients:
\begin{equation}
\label{eq:cor}
    r_{\mathcal{M}_{t},\ \mathcal{P}} = (r_{\mathcal{M}_{t},\ \mathcal{P}}^{Pearson} + r_{\mathcal{M}_{t},\mathcal{P}}^{Spearman})/{2}
\end{equation}
Since our experiments minimize confounding factors, variations in model performance can be more directly attributed to differences in IT dataset diversity. Thus, the correlation $r_{\mathcal{M}_{t},\ \mathcal{P}}$ indicates how reliably each metric captures "IT-aligned Diversity"\footnote{Unless otherwise specified, the term "diversity" in this paper generally refers to "IT-aligned Diversity."}—the type of data diversity beneficial for instruction tuning LLMs.

The results are shown in Figure \ref{fig:exist}, with supplementary plots in Appendix \ref{app:results}. Overall, we observe that each metric tends to favor datasets aligned with its own selection criterion, but may not correlate strongly with performance due to overlooking other aspects of diversity:

\paragraph{Findings 1}
\textit{Lexical diversity metrics fail to distinguish between different samples and datasets, showing weak correlation with model performance.}

As shown in Figure \ref{fig:exist}(a, b), high- and low-performance datasets exhibit similar lexical diversity. This likely results from the widespread use of diverse vocabulary in IT samples, making lexical diversity an ineffective measure for IT datasets.

\paragraph{Findings 2}
\textit{Since distribution-based semantic diversity metrics neglect sample uniqueness, they often underestimate the diversity of datasets with large inter-sample distances.}

From Figure \ref{fig:exist}(c, d), we observe that datasets selected by Farthest and K-Center-Greedy (purple and blue points) achieve high IT performance but often receive relatively lower diversity scores from distribution-based diversity metrics, thus weakening their correlation with model performance. This likely occurs because these strategies all prioritize sample uniqueness by selecting samples that are distant from others, a factor not captured by distribution-based metrics. This suggests that overlooking sample uniqueness diminishes the reliability of diversity metrics.

\paragraph{Findings 3}
\textit{As distance-based semantic diversity metrics neglect information density in semantic space, they often underestimates datasets that are close to the overall sample distribution and overestimates datasets with large inter-sample distances.}

From Figure \ref{fig:exist}(e, f, g, h), we observe common outliers in the fitting line for datasets selected by QDIT and K-means (orange and green points), which receive low diversity scores despite strong performance according to distance-based diversity metrics. In contrast, K-Center-Greedy and Repr Filter (blue and red points) show the opposite trend, weakening the metrics' correlation with the model performance. This is likely because the former two strategies select more samples from dense semantic regions, which better cover the overall sample distribution but conflicts with distance-based diversity calculations. This suggests that ignoring information density in semantic space reduces the reliability of diversity metrics.

\paragraph{Findings 4}
\textit{Distance-based metrics often fail to accurately measure diversity in datasets containing redundant samples.}

As shown by the duplicated datasets (gray points) in Figure \ref{fig:exist}(e, f, g, h), DistSum fails to capture redundancy effectively, as total distances are dominated by variations in distant samples. Meanwhile, other metrics, such as KNN Distance, overly penalize redundant samples by nullifying their contribution to overall diversity.

%% file: latex/sections/novelsum.tex
\section{Proposed Metric: \textit{NovelSum}}
\label{sec:PM}
Extending previous findings, we derive some insights on how to design a more reliable metric: (1) \textbf{The uniqueness of individual samples should be a key factor in measuring dataset diversity.} This uniqueness stems from sufficient inter-sample distances, providing diverse information that helps the model learn more generalized patterns. (2) \textbf{When quantifying a sample's uniqueness, its distance to nearby and distant samples should be balanced.} Differences with nearby samples define uniqueness and should hold greater importance, with weights assigned smoothly. (3) \textbf{When calculating inter-sample distances, both semantic differences and local information density should be considered.} In practical applications of instruction fine-tuning, semantic space varies in information density, with scenarios like math and code having denser data and information. Focusing only on semantics overlooks valuable fine-grained information for the model.

\begin{figure}[t!]
    \centering
        \includegraphics[width=\linewidth]{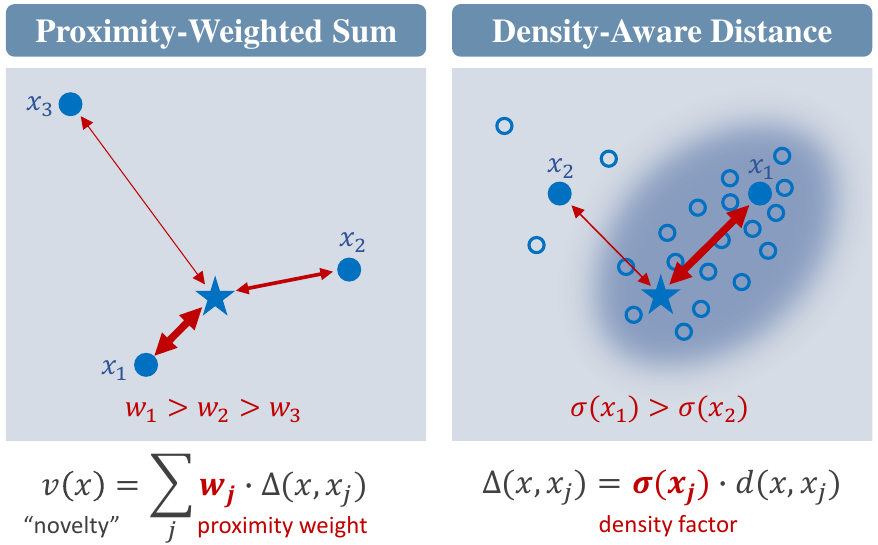}
    \caption{\textit{NovelSum} computes each sample's novelty as a proximity-weighted sum of its density-aware distances to other samples, where closer points have greater influence and high-density regions produce larger distances.}
    \label{fig:novelsum}
    \vspace{-4mm}
\end{figure}

Following these principles, we introduce \textit{NovelSum}, a diversity metric that jointly considers distance and distribution. Specifically, we define dataset diversity as the sum of each sample's uniqueness---its unique contribution to overall information, which we later term "novelty":
\begin{equation}
\label{eq:def}
    \mathcal{M}_{NovelSum}(\mathcal{X}) = \sum_{x_i \in \mathcal{X}} v(x_i)
\end{equation}
Figure \ref{fig:novelsum} and the following paragraphs illustrate how each sample's novelty is computed.

\paragraph{Proximity-Weighted Sum}
In contrast to DistSum (Eq.~\ref{eq:distsum}), which calculates a sample’s uniqueness as a simple sum of distances to other points, we propose a proximity-weighted sum that assigns higher weights to closer points, giving them a larger influence on the uniqueness score:
\begin{equation}
\label{eq:pws}
    v(x_i) = \sum_{x_j \in \mathcal{X},\ x_j \neq x_i} w(x_i, x_j)^{\alpha} \cdot \Delta(x_i, x_j),
\end{equation}
where the proximity weight is defined as:
\begin{equation*}
    w(x_i, x_j) = \phi(\pi_i(j))
\end{equation*}
Here, \( \pi_i(j) \) is the rank of \( x_j \) in the sorted list of distances from \( x_i \) to all other points in \( \mathcal{X} \), with \( \pi_i(j) = 1 \) indicating that \( x_j \) is the nearest neighbor of \( x_i \). The function \( \phi(\cdot) \) is monotonically decreasing, smoothing the weights according to the proximity, for example, we set \( \phi(\pi_i(j)) = \frac{1}{\pi_i(j)} \). The hyperparameter \( \alpha \) controls the degree to which proximity impacts the uniqueness score.

\paragraph{Density-Aware Distance}
To account for the local information density when calculating \( \Delta(x_i, x_j) \), we introduce a density-aware distance that multiplies the original semantic distance by a density factor \( \sigma(x_j) \):
\begin{equation}
\label{eq:dad}
    \Delta(x_i, x_j) = \sigma(x_j)^{\beta} \cdot d(x_i, x_j)
\end{equation}
Since the probabilistic density of the overall sample distribution is intractable, we approximate the density factor by the inverse of the average distance to the \( K \)-nearest neighbors of \( x_j \) in \( \mathcal{X}^{all} \):
\begin{equation*}
    \sigma(x_j) = \frac{1}{\sum_{k=1}^{K} d(x_j, N_k(x_j))}
\end{equation*}
Here, \( d(\cdot, \cdot) \) represents the distance between the embeddings of samples (e.g., cosine distance), and \( N_k(x) \) denotes the \( k \)-th nearest neighbor of \( x \). The hyperparameter \( \beta \) controls the extent to which density influences the distance. The reference dataset \( \mathcal{X}^{all} \) can be replaced to estimate information density under different sample distributions.

This approach mirrors how novelty is assessed in academic papers: a paper's novelty lies in its difference from closely related work, measured within the context of its field for greater accuracy. Accordingly, we treat each sample's quantified uniqueness as its "novelty" and name our method "NovelSum."
Further details are provided in Appendix~\ref{app:impl_novelsum} (implementation), Appendix~\ref{app:theo} (theoretical interpretation), and Appendix~\ref{app:complexity} (computational complexity analysis highlighting \textit{NovelSum}'s efficiency).

%% file: latex/sections/simulation.tex
\begin{figure*}[!t]
    \centering
        \includegraphics[width=\linewidth]{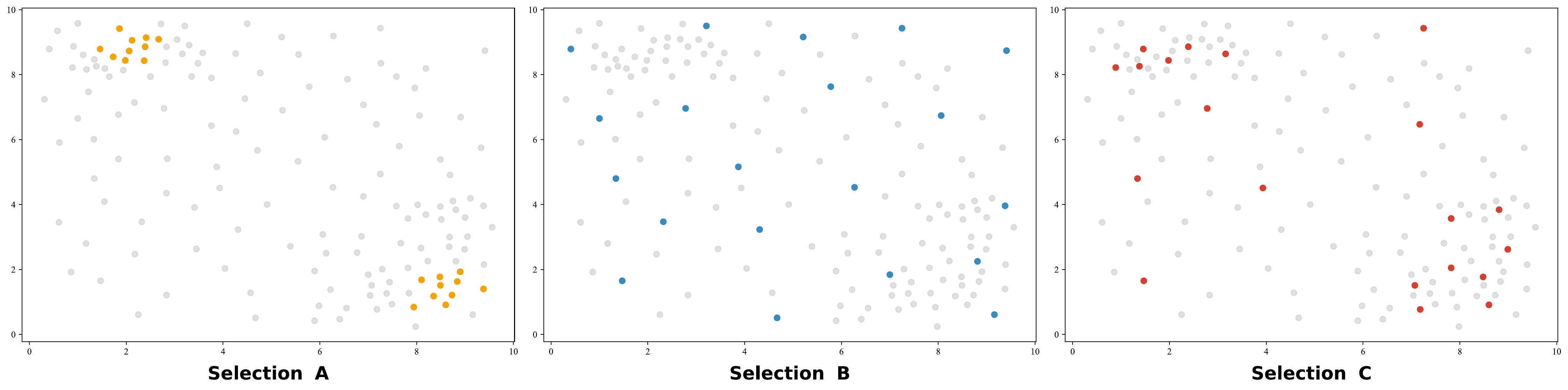}
    \caption{Simulating different data selection scenarios in a 2D sample space: Selection A represents datasets with redundancy, Selection B optimizes inter-sample distances, and Selection C accounts for both distances and density, which prior analysis suggests yields the highest diversity for instruction tuning.}
    \label{fig:sim_points}
    \vspace{-4mm}
\end{figure*}

\begin{figure}[!t]
    \centering
        \includegraphics[width=\linewidth]{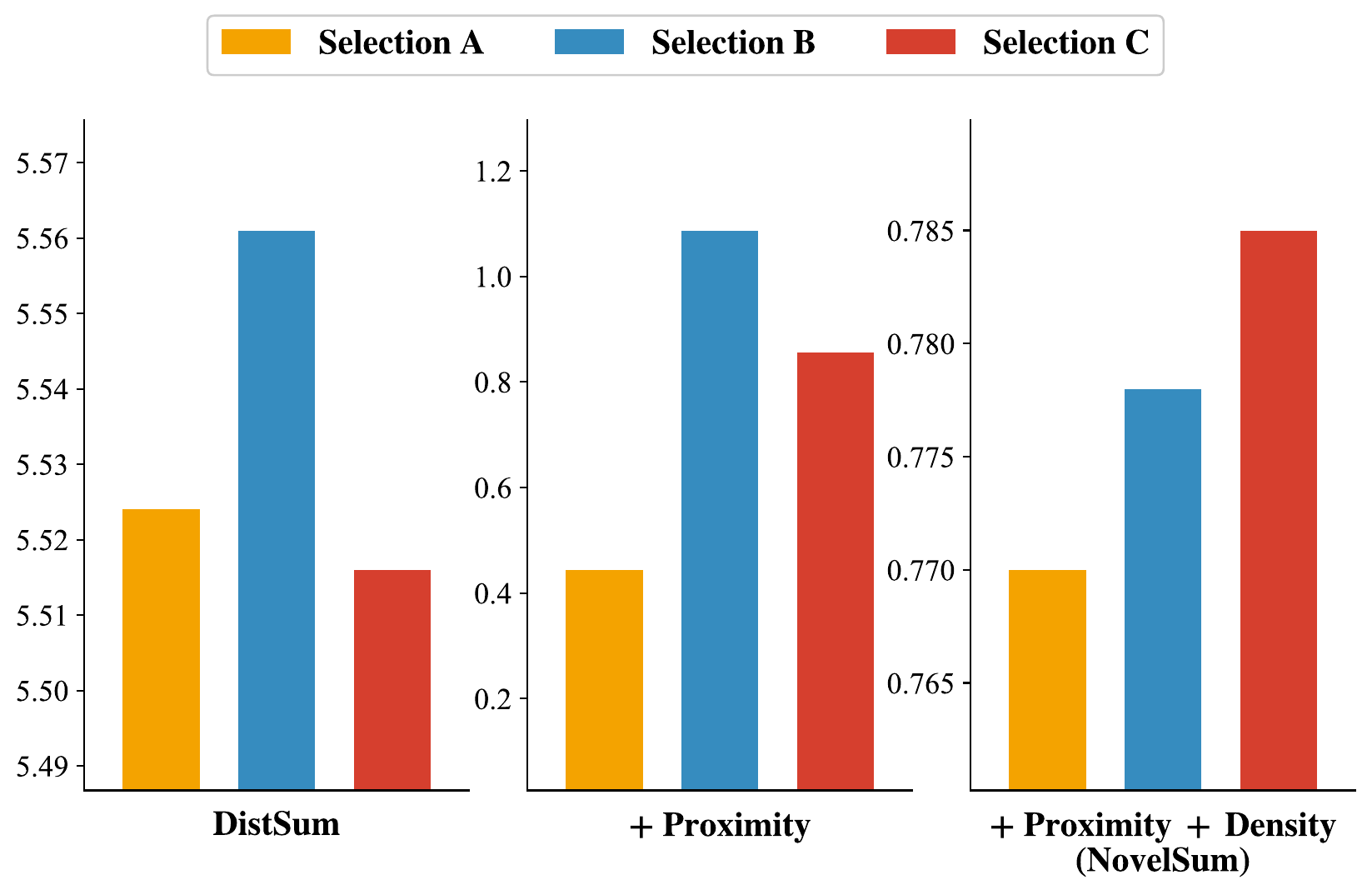}
    \caption{Measuring the diversity of simulated selection A/B/C with various metrics. \textit{NovelSum} accurately captures dataset diversity, exhibiting expected behaviors.}
    \label{fig:sim_res}
    \vspace{-4mm}
\end{figure}

\section{Simulation Study}

To validate whether the proposed metric aligns with our design principles and accurately captures dataset diversity, we create a visualizable simulation environment. We generate 150 points in 2D space as the data source and select 20 samples to form a dataset, simulating the data selection process for instruction tuning. As shown in Figure \ref{fig:sim_points}, we analyze three data selection scenarios to examine the behavior of our diversity metric. "Selection A" contains samples from two clusters, with most points close to each other, simulating datasets with redundancy. "Selection B", constructed using K-Center-Greedy, consists of samples far apart, simulating datasets optimized for inter-sample semantic distances. "Selection C" considers both inter-sample distances and information density, simulating datasets that best represent the sample space with unique points. Based on prior analysis, the dataset diversity of the three selections is expected to follow $A<B<C$ order under the IT scenario.

Figure \ref{fig:sim_res} presents the diversity measurement results using DistSum, a proximity-weighted version of DistSum, and \textit{NovelSum}. From left to right, we see that DistSum counterintuitively considers $\mathcal{M}(A) \simeq \mathcal{M}(C)$, failing to reflect sample uniqueness. Incorporating the proximity-weighted sum improves uniqueness capture but still exhibits $\mathcal{M}(B) > \mathcal{M}(C)$, overlooking information density. \textbf{\textit{NovelSum} resolves these issues, accurately capturing diversity variations in alignment with design principles}, yielding $\mathcal{M}(A) < \mathcal{M}(B) < \mathcal{M}(C)$. This study further validates the necessity of the proximity-weighted sum and density-aware distance for precise diversity measurement.

%% file: latex/sections/experiment.tex
\section{Experiments}
\label{sec:exp}
Following the settings in Section \ref{sec:existing}, we evaluate \textit{NovelSum}'s correlation with the fine-tuned model performance across 53 IT datasets and compare it with previous diversity metrics. Additionally, we conduct a correlation analysis using Qwen-2.5-7B \cite{yang2024qwen2} as the backbone model, alongside previous LLaMA-3-8B experiments, to further demonstrate the metric's effectiveness across different scenarios. 
Due to resource constraints, we run each strategy on Qwen for at least two rounds, yielding 25 datasets. 

\subsection{Main Results}

\begin{table*}[!t]
    \centering
    \resizebox{\linewidth}{!}{
    \begin{tabular}{lcccccccccc}
    \toprule
    \multirow{3}*{\textbf{Metrics}} & \multicolumn{10}{c}{\textbf{Data Selection Strategies}} \\
    \cmidrule(lr){2-11}
    & \multirow{2}*{\textbf{K-means}} & \scalebox{0.95}{\multirow{2}*{\vtop{\hbox{\textbf{K-Center}}\vspace{1mm}\hbox{\textbf{-Greedy}}}}}  & \multirow{2}*{\textbf{QDIT}} & \multirow{2}*{\vtop{\hbox{\textbf{Repr}}\vspace{1mm}\hbox{\textbf{Filter}}}} & \multicolumn{5}{c}{\textbf{Random}} & \scalebox{0.95}{\multirow{2}{*}{\textbf{Duplicate}}} \\ 
    \cmidrule(lr){6-10}
    & & & & & \textbf{$\mathcal{X}^{all}$} & \scalebox{0.95}{ShareGPT} & \scalebox{0.95}{WizardLM} & Alpaca & Dolly &  \\
    \midrule
    \rowcolor{gray!15} \multicolumn{11}{c}{\textit{LLaMA-3-8B}} \\
    \noalign{\vskip 1mm}
    \textbf{Model Performance} & \cellcolor{BLUE!60}1.32 & \cellcolor{BLUE!50}1.31 & \cellcolor{BLUE!40}1.25 & \cellcolor{BLUE!30}1.05 & \cellcolor{BLUE!20}1.20 & \cellcolor{BLUE!10}0.83 & \cellcolor{BLUE!0}0.72 & \cellcolor{ORANGE!10}0.07 & \cellcolor{ORANGE!20}-0.14 & \cellcolor{ORANGE!30}-1.35 \\
    \midrule    
    \textbf{NovelSum (Ours)} & \cellcolor{BLUE!60} 0.693 & \cellcolor{BLUE!50} 0.687 & \cellcolor{BLUE!30} 0.673 & \cellcolor{BLUE!20} 0.671 & \cellcolor{BLUE!40} 0.675 & \cellcolor{BLUE!10} 0.628 & \cellcolor{BLUE!0} 0.591 & \cellcolor{ORANGE!10} 0.572 & \cellcolor{ORANGE!20} 0.50 & \cellcolor{ORANGE!30} 0.461 \\
    Vendi Score $_{\times10^7}$ & \cellcolor{BLUE!30} 1.70 & \cellcolor{BLUE!60} 2.53 & \cellcolor{BLUE!10} 1.59 & \cellcolor{BLUE!50} 2.23 & \cellcolor{BLUE!20} 1.61 & \cellcolor{BLUE!30} 1.70 & \cellcolor{ORANGE!10} 1.44 & \cellcolor{ORANGE!20} 1.32 & \cellcolor{ORANGE!10} 1.44 & \cellcolor{ORANGE!30} 0.05 \\
    DistSum$_{cosine}$  & \cellcolor{BLUE!30} 0.648 & \cellcolor{BLUE!60} 0.746 & \cellcolor{BLUE!0} 0.629 & \cellcolor{BLUE!50} 0.703 & \cellcolor{BLUE!10} 0.634 & \cellcolor{BLUE!40} 0.656 & \cellcolor{ORANGE!30} 0.578 & \cellcolor{ORANGE!10} 0.605 & \cellcolor{ORANGE!20} 0.603 & \cellcolor{BLUE!10} 0.634 \\
    Facility Loc. $_{\times10^5}$ & \cellcolor{BLUE!40} 2.99 & \cellcolor{ORANGE!10} 2.73 & \cellcolor{BLUE!40} 2.99 & \cellcolor{BLUE!20} 2.86 & \cellcolor{BLUE!40} 2.99 & \cellcolor{BLUE!0} 2.83 & \cellcolor{BLUE!30} 2.88 & \cellcolor{BLUE!0} 2.83 & \cellcolor{ORANGE!20} 2.59 & \cellcolor{ORANGE!30} 2.52 \\    
    \midrule
    \midrule
    \rowcolor{gray!15} \multicolumn{11}{c}{\textit{Qwen-2.5-7B}} \\
    \noalign{\vskip 1mm}
    \textbf{Model Performance} & \cellcolor{BLUE!30} 1.06 & \cellcolor{BLUE!60} 1.45 & \cellcolor{BLUE!40} 1.23 & \cellcolor{BLUE!50} 1.35 & \cellcolor{BLUE!20} 0.87 & \cellcolor{BLUE!10} 0.07 & \cellcolor{BLUE!0} -0.08 & \cellcolor{ORANGE!10} -0.38 & \cellcolor{ORANGE!30} -0.49 & \cellcolor{ORANGE!20} -0.43 \\
    \midrule
    \textbf{NovelSum (Ours)}  & \cellcolor{BLUE!40} 0.440 & \cellcolor{BLUE!60} 0.505 & \cellcolor{BLUE!20} 0.403 & \cellcolor{BLUE!50} 0.495 & \cellcolor{BLUE!30} 0.408 & \cellcolor{BLUE!10} 0.392 & \cellcolor{BLUE!0} 0.349 & \cellcolor{ORANGE!10} 0.336 & \cellcolor{ORANGE!20} 0.320 & \cellcolor{ORANGE!30} 0.309 \\
    Vendi Score $_{\times10^6}$ & \cellcolor{ORANGE!10} 1.60 & \cellcolor{BLUE!40} 3.09 & \cellcolor{BLUE!10} 2.60 & \cellcolor{BLUE!60} 7.15 & \cellcolor{ORANGE!20} 1.41 & \cellcolor{BLUE!50} 3.36 & \cellcolor{BLUE!20} 2.65 & \cellcolor{BLUE!0} 1.89 & \cellcolor{BLUE!30} 3.04 & \cellcolor{ORANGE!30} 0.20 \\
    DistSum$_{cosine}$ & \cellcolor{BLUE!30} 0.260 & \cellcolor{BLUE!60} 0.440 & \cellcolor{BLUE!0} 0.223 & \cellcolor{BLUE!50} 0.421 & \cellcolor{BLUE!10} 0.230 & \cellcolor{BLUE!40} 0.285 & \cellcolor{ORANGE!20} 0.211 & \cellcolor{ORANGE!30} 0.189 & \cellcolor{ORANGE!10} 0.221 & \cellcolor{BLUE!20} 0.243 \\
    Facility Loc. $_{\times10^5}$ & \cellcolor{BLUE!40} 3.54 & \cellcolor{ORANGE!30} 3.42 & \cellcolor{BLUE!40} 3.54 & \cellcolor{ORANGE!20} 3.46 & \cellcolor{BLUE!40} 3.54 & \cellcolor{BLUE!30} 3.51 & \cellcolor{BLUE!10} 3.50 & \cellcolor{BLUE!10} 3.50 & \cellcolor{ORANGE!20} 3.46 & \cellcolor{BLUE!0} 3.48 \\ 
    \bottomrule
    \end{tabular}
    }
    \caption{Comparison of fine-tuned model performance (Eq.~\ref{eq:perf}) and diversity measurement results across datasets selected by different strategies. Each row visualizes the relative ranking of metric scores across datasets using color intensity: darker \colorbox{BLUE!60}{blue} indicates higher values per row, and darker \colorbox{ORANGE!30}{orange} indicates lower values. \textit{NovelSum} consistently shows a stronger correlation with model performance than other metrics, even as data selection strategies vary in performance between LLaMA-3-8B and Qwen-2.5-7B. More results are provided in Appendix \ref{app:results}.
    }
    \label{tbl:main}
    \vspace{-4mm}
\end{table*}

\begin{table}[t!]
\centering
\resizebox{\linewidth}{!}{
\begin{tabular}{lcccc}
\toprule
\multirow{2}*{\textbf{Diversity Metrics}} & \multicolumn{3}{c}{\textbf{LLaMA}} & \textbf{Qwen}\\
\cmidrule(lr){2-4} \cmidrule(lr){5-5} 
& \textbf{Pearson} & \textbf{Spearman} & \textbf{Avg.} & \textbf{Avg.} \\
\midrule
TTR & -0.38 & -0.16 & -0.27 & -0.30 \\
vocd-D & -0.43 & -0.17 & -0.30 & -0.31 \\
\midrule
Facility Loc. & 0.86 & 0.69 & 0.77 & 0.08 \\
Entropy & 0.93 & 0.80 & 0.86 & 0.63 \\
\midrule
LDD & 0.61 & 0.75 & 0.68 & 0.60 \\
KNN Distance & 0.59 & 0.80 & 0.70 & 0.67 \\
DistSum$_{cosine}$ & 0.85 & 0.67 & 0.76 & 0.51 \\
Vendi Score & 0.70 & 0.85 & 0.78 & 0.60 \\
DistSum$_{L2}$ & 0.86 & 0.76 & 0.81 & 0.51 \\
Cluster Inertia & 0.81 & 0.85 & 0.83 & 0.76 \\
Radius & 0.87 & 0.81 & 0.84 & 0.48 \\
\midrule
NovelSum & \textbf{0.98} & \textbf{0.95} & \textbf{0.97} & \textbf{0.90} \\
\bottomrule
\end{tabular}
}
\caption{Correlations between different metrics and model performance on LLaMA-3-8B and Qwen-2.5-7B.  “Avg.” denotes the average correlation (Eq. \ref{eq:cor}).}
\label{tbl:correlations}
\vspace{-4mm}
\end{table}

\paragraph{\textit{NovelSum} consistently achieves state-of-the-art correlation with model performance across various data selection strategies, backbone LLMs, and correlation measures.}
Table \ref{tbl:main} presents diversity measurement results on datasets constructed by mainstream data selection methods (based on $\mathcal{X}^{all}$), random selection from various sources, and duplicated samples (with only $m=100$ unique samples). 
Results from multiple runs are averaged for each strategy.
Although these strategies yield varying performance rankings across base models, \textit{NovelSum} consistently tracks changes in model performance by accurately measuring dataset diversity. For instance, K-means achieves the best performance on LLaMA with the highest NovelSum score, while K-Center-Greedy excels on Qwen, also correlating with the highest NovelSum. Table \ref{tbl:correlations} shows the correlation coefficients between various metrics and fine-tuned model performance for both LLaMA and Qwen experiments, where \textit{NovelSum} achieves state-of-the-art correlation across different models and measures.

\paragraph{\textit{NovelSum} can provide valuable guidance for data engineering practices.}
As a reliable indicator of data diversity, \textit{NovelSum} can assess diversity at both the dataset and sample levels, directly guiding data selection and construction decisions. For example, Table \ref{tbl:main} shows that the combined data source $\mathcal{X}^{all}$ is a better choice for sampling diverse IT data than other sources. Moreover, \textit{NovelSum} can offer insights through comparative analyses, such as: (1) ShareGPT, which collects data from real internet users, exhibits greater diversity than Dolly, which relies on company employees, suggesting that IT samples from diverse sources enhance dataset diversity \cite{wang2024diversity-logD}; (2) In LLaMA experiments, random selection can outperform some mainstream strategies, aligning with prior work \cite{xia2024rethinking,diddee2024chasing}, highlighting gaps in current data selection methods for optimizing diversity.

\subsection{Ablation Study}

\textit{NovelSum} comprises several tunable components. In our main experiments, we use cosine distance to compute $d(x_i, x_j)$ in Eq.~\ref{eq:dad}, with hyperparameters set to $\alpha = 1$, $\beta = 0.5$, and $K = 10$ nearest neighbors in Eq.~\ref{eq:pws} and Eq.~\ref{eq:dad}. Here, we conduct an ablation study to investigate the impact of these settings based on LLaMA-3-8B.

\begin{table}[t!]
\centering
\resizebox{\linewidth}{!}{
\begin{tabular}{lccc}
\toprule
\textbf{Variants} & \textbf{Pearson} & \textbf{Spearman} & \textbf{Avg.} \\
\midrule
NovelSum & 0.98 & 0.95 & 0.97 \\
\midrule
\hspace{0.10cm} - Use $L2$ distance & 0.97 & 0.83 & 0.90\textsubscript{↓ 0.08} \\
\hspace{0.10cm} - $K=20$ & 0.98 & 0.96 & 0.97\textsubscript{↓ 0.00} \\
\hspace{0.10cm} - $\alpha=0$ (w/o proximity) & 0.79 & 0.31 & 0.55\textsubscript{↓ 0.42} \\
\hspace{0.10cm} - $\beta=0$ (w/o density) & 0.92 & 0.89 & 0.91\textsubscript{↓ 0.07} \\
\bottomrule
\end{tabular}
}
\caption{Ablation Study for \textit{NovelSum}.}
\label{tbl:ablation}
\vspace{-4mm}
\end{table}

In Table \ref{tbl:ablation}, $\alpha=0$ removes the proximity weights, and $\beta=0$ eliminates the density multiplier. We observe that both $\alpha=0$ and $\beta=0$ significantly weaken the correlation, validating the benefits of the proximity-weighted sum and density-aware distance. 
Replacing cosine distance with Euclidean distance and using more neighbors for density approximation have minimal impact, particularly on Pearson's correlation, highlighting \textit{NovelSum}'s robustness to different distance measures.
Additionally, Appendix~\ref{app:hyper} presents an in-depth analysis of the hyperparameters, demonstrating the reliability of our current configurations and providing guidance for broader application.

%% file: latex/sections/novelselect.tex
\section{Data Selection Strategy}
\label{sec:DSS}

\paragraph{Introducing \textit{NovelSelect}}

Given \textit{NovelSum}'s accurate diversity measurement and strong correlation with model performance, we investigate its potential as an optimization objective for selecting samples and generating a diverse dataset:
\begin{equation}
\label{eq:argmax}
    \mathcal{X} = \arg\max_{\mathcal{X} \subset \mathcal{X}^{all}} \mathcal{M}_{NovelSum}(\mathcal{X}),
\end{equation}
where $\mathcal{M}_{NovelSum}(\mathcal{X})$ is defined in Eq.~\ref{eq:def}. Since directly solving Eq.~\ref{eq:argmax} is NP-hard \cite{cook1994combinatorial}, we propose a greedy approach that iteratively selects the most "novel" sample to maximize the \textit{NovelSum} of the final dataset. The "novelty" of a new sample $x$ relative to an existing set $\mathcal{X}$ is defined as:
\begin{equation}
\label{eq:v}
    v(x) = \sum_{x_j \in \mathcal{X}} w(x, x_j)^{\alpha} \cdot\sigma(x_j)^{\beta} \cdot d(x, x_j),
\end{equation}
where $w(x, x_j)$ and $\sigma(x_j)$ are the proximity weight and density factor from Eq. \ref{eq:pws} and \ref{eq:dad}. At each step, the sample with the highest novelty is selected: $x^{new} = \arg\max_{x}v(x), \quad \mathcal{X} \leftarrow {x^{new}} \cup \mathcal{X}$. This process is repeated from $\mathcal{X} = \emptyset$ until the data budget is reached, resulting in the selected dataset. We refer to this approach as \textit{NovelSelect}. 

\input{latex/table/algorithm}

Algorithm \ref{alg:novelselect} outlines the overall process with additional details provided in Appendix~\ref{app:impl_novelselect}. Notably, \textit{NovelSelect} is as computationally efficient as existing approaches, with a detailed analysis provided in Appendix~\ref{app:complexity}. Furthermore, by incorporating quality scores into $v(x)$, \textit{NovelSelect} can seamlessly integrate with quality-based data selection methods, highlighting its extensibility.

\paragraph{Data Selection Experiments}

We conduct additional data selection experiments on LLaMA-3-8B to evaluate \textit{NovelSelect}'s performance. Following prior settings, we use \textit{NovelSelect} to select 10k samples from $\mathcal{X}^{all}$ and assess the fine-tuned model's performance on MT-bench and AlpacaEval. Results are averaged over three runs. 

\begin{table}[t!]
\centering
\resizebox{\linewidth}{!}{
\begin{tabular}{lcc|c}
\toprule
\textbf{Strategies} & \textbf{MT-bench} & \textbf{AlpacaEval} & \textbf{Aggregated $\mathcal{P}$} \\
\midrule
Random & 6.18 & 75.47 & 1.20 \\
Repr Filter & 6.17 & 72.57 & 1.05 \\
QDIT & 6.21 & 75.91 & 1.25 \\
K-Center-Greedy & 6.33 & 75.30 & 1.31 \\
K-means & 6.33 & 75.46 & 1.32 \\
\midrule
NovelSelect & \textbf{6.47} & \textbf{78.07} & \textbf{1.55} \\
\bottomrule
\end{tabular}
}
\caption{Comparisons of different diversity-oriented data selection strategies on IT performance. $\mathcal{P}$ aggregates the performance based on Z-scores (Eq. \ref{eq:perf}).}
\label{tbl:select}
\vspace{-4mm}
\end{table}

From Table \ref{tbl:select}, \textit{NovelSelect} outperforms existing diversity-oriented data selection strategies on both benchmarks, demonstrating superior IT performance. This aligns with the higher \textit{NovelSum} scores achieved by \textit{NovelSelect} (Figure~\ref{fig:head}), further validating \textit{NovelSum}'s effectiveness and practical value in IT data engineering.

%% file: latex/table/algorithm.tex
\begin{algorithm}[t!]
\caption{\textit{NovelSelect}}
\label{alg:novelselect}
\begin{algorithmic}[1]
\State \textbf{Input:} Data pool $\mathcal{X}^{all}$, data budget $n$
\State Initialize an empty dataset, $\mathcal{X} \gets \emptyset$
\While{$|\mathcal{X}| < n$}
    \State $x^{new} \gets \arg\max_{x \in \mathcal{X}^{all}} v(x)$
    \State $\mathcal{X} \gets \mathcal{X} \cup \{x^{new}\}$
    \State $\mathcal{X}^{all} \gets \mathcal{X}^{all} \setminus \{x^{new}\}$
\EndWhile
\State \textbf{return} $\mathcal{X}$
\end{algorithmic}
\end{algorithm}

%% file: latex/sections/discussion.tex
\section{Discussion}

\paragraph{From General IT to Downstream Tasks}
Our study focuses on general instruction tuning, with training and evaluation covering a wide range of downstream tasks, thereby offering insights applicable to broader real-world scenarios.
At the same time, extending performance-aligned diversity measurement to specific domains—such as math or code—is also valuable and may warrant dedicated investigation. A promising approach is to adapt \textit{NovelSum} by simply replacing the general dataset $\mathcal{X}^{all}$ used in density estimation with domain-specific data sources. We hope our work lays a solid foundation for future research in this direction.

\paragraph{Impact of Embedding Extractor}

Different embedding extractors may yield different sample distributions in the semantic space, potentially affecting diversity computations. In our study, we use LLaMA-3-8B to compute embeddings for experiments based on LLaMA-3-8B, and Qwen-2.5-7B for those based on Qwen-2.5-7B (see Appendix~\ref{app:preprocess} for embedding details). This setup is motivated by an interesting observation: extracting embeddings using the same model as the fine-tuning backbone yields metrics with the highest correlation to instruction tuning performance. The corresponding experiment is described in detail in Appendix~\ref{app:embed}.

%% file: latex/sections/related.tex
\section{Related Work}
\paragraph{Measuring Dataset Diversity}

Dataset diversity is essential for training generalizable machine learning models, drawing significant research interest \cite{tevet2021evaluating, sun2024diversity-CV, zhang2024onlyif, zhao2024measuring, qin2024unleashing}. In NLP, numerous lexical diversity metrics have been proposed to measure text diversity through vocabulary usage \cite{richards1987type-TTR, malvern2004lexical-vocd}. Recently, semantic embeddings have enabled more flexible diversity measurement from a distance-based perspective \cite{du2019boosting-Inertia, yu2022can, stasaski2022semantic-KNN, dang2024data} or a distribution-based perspective \cite{han2022measuring, shao2024balanced}. Focusing on instruction tuning, while some studies have explored the assessment of IT data diversity \cite{wang2024diversity-logD, bukharin2023data-QDIT}, the proposed metrics lack sufficient validation of their correlation with IT performance; thus, reliable metrics for guiding data engineering remain underexplored.

\paragraph{Data Selection for Instruction Tuning}
Instruction tuning trains LLMs to follow human instructions using instruction-response pairs \cite{zhang2023instruction}. While earlier work focused on large-scale IT datasets \cite{longpre2023flan, chiang2023vicuna-ShareGPT}, recent studies show that small, high-quality data sets can reduce costs and improve performance \cite{chen2023maybe-Kcentergreedy, chen2023alpagasus, zhou2024lima, dou2024loramoe, ye2024empirical}. This has led to the development of data selection strategies to identify subsets that boost IT performance, including many effective diversity-aware approaches \cite{liu2023makes, du2023mods-Kcentergreedy, wu2023self-Kcentergreedy, song2024iterselecttune-Kmeans, ge2024clustering, kung2023active, yang2024beyond, yang2025diversity}. However, the lack of clear definitions and reliable diversity metrics for IT datasets hinders effective optimization. Consequently, some selection methods fail to generalize or or even underperform random selection \cite{xia2024rethinking,diddee2024chasing}. Addressing this gap, our work seeks to provide a more reliable diversity metric, grounded in comprehensive analysis, that accurately captures dataset diversity and better aligns with instruction tuning performance, thus offering practical guidance for data selection.

%% file: latex/sections/appendix.tex
\section{Details of Correlation Evaluation}

\subsection{Data Processing and Semantic Embeddings}
\label{app:preprocess}
We apply basic preprocessing to remove anomalous samples from the data sources, ensuring more stable results while preserving generality. In the early stage of our work, we observe that short samples often exhibit low quality and tend to be outliers in the semantic space, potentially distorting experimental results. To address this, we filter out samples shorter than 256 tokens using the BERT \cite{devlin2018bert} tokenizer, ensuring consistency for experiments across different LLMs. Furthermore, to ensure the dataset's relevance for English-language tasks and math problems, we exclude samples with a non-English-or-number ratio exceeding 0.8. 

When computing sample embeddings, we set the maximum sequence length to 256 to mitigate bias from varying text lengths. This applies only to embedding computation; fine-tuning uses a much larger maximum length. We extract the last hidden layer of the language model and apply mean pooling, excluding padding tokens, to generate robust sample-level embeddings. We analyze and discuss the choice of embedding extractors in Appendix~\ref{app:embed}. Based on this analysis, we use LLaMA-3-8B to compute embeddings in experiments using LLaMA-3-8B as the fine-tuning backbone. Similarly, for experiments using Qwen-2.5-7B as the fine-tuning backbone, we use Qwen-2.5-7B to compute embeddings.

\subsection{Details of Existing Diversity Metrics}
\label{app:exist}

For lexical diversity, the \textbf{Type-Token Ratio} (TTR) quantifies the lexical diversity of a text sequence $x_i$ as the ratio of distinct tokens to the total number of tokens. The overall lexical diversity of a dataset $\mathcal{X} = \{x_1, x_2, ..., x_N\}$ is computed as the average TTR across all samples:  
\begin{equation}  
    \mathcal{M}_{TTR}(\mathcal{X}) = \frac{1}{N} \sum_{i=1}^{N} \frac{|Unique(x_i)|}{|x_i|}.  
\end{equation}  
To mitigate the influence of text length on TTR, we randomly sample 30 tokens from each data point to compute the TTR.  

To address the sensitivity of TTR to text length, \textbf{vocd-D} extends this measure by computing $TTR_i^k$ over sampled sub-sequences of varying lengths $k$ and fitting the following curve:  
\begin{equation}  
    \hat{TTR}_i^k = \frac{D}{k} \left( (1 + 2 \frac{k}{D})^{\frac{1}{2}} - 1 \right),  
\end{equation}  
where $D$ is the estimated parameter representing lexical diversity. The vocd-D metric is defined as $\mathcal{M}_{vocd-D} = D_{\text{best fit}}$, with larger values indicating greater lexical diversity. In our experiments, we compute $TTR_i^k$ for $k = 10, 20, 30, 40, 50$ and take the average of the resulting values as the final lexical diversity score. 

For distance-based semantic diversity, 
\textbf{Cluster Inertia} \cite{du2019boosting-Inertia} quantifies diversity by partitioning the dataset into $K$ clusters using K-means and summing the squared distances between each sample and its cluster centroid:  
\begin{equation}  
    \mathcal{M}_{Inertia}(\mathcal{X}) = \sum_{j=1}^{K} \sum_{x_i \in C_j} \|emb(x_i) - \mu_j\|^2,  
\end{equation}  
where $\mu_j$ is the centroid of cluster $C_j$. Higher inertia indicates greater sample dispersion. In practice, we cluster $\mathcal{X}$ into 200 clusters for subsequent computations.
Moreover, \textbf{Vendi Score} (VS) \cite{pasarkar2023cousins-Vendi} measures diversity based on the eigenvalues of the similarity kernel matrix. The generalized VS metric is defined as:  
\begin{equation}  
    \mathcal{M}_{VS}(\mathcal{X}) = \exp\left(\frac{1}{1 - \alpha} \log_2 \sum_{i=1}^{|\mathcal{X}|} \bar{\lambda}_{i|\theta}^{\alpha} \right),  
\end{equation}  
where $\bar{\lambda}_{i|\theta}$ represents the normalized eigenvalues. We set $\alpha=0.5$ to enhance measurement under severe class imbalance.
\textbf{Radius} \cite{lai2020diversity-Radius} characterizes the dispersion of the sample space by approximating embeddings as a multi-variate Gaussian distribution. It computes the geometric mean of the standard deviations along each dimension:  
\begin{equation}  
    \mathcal{M}_{Radius}(\mathcal{X}) = \sqrt[H]{\prod_{j=1}^{H} \sigma_j},  
\end{equation}  
where $H$ is the embedding dimension, and $\sigma_j$ denotes the radius of the ellipsoid along the $j$-th axis. Larger values indicate a greater spread of samples in the embedding space. \textbf{Log Determinant Distance} \cite{wang2024diversity-logD} utilizes the determinant of the similarity matrix as a measure of dataset diversity. In our work, we employ the cosine similarity function to compute the similarity matrix. 

For \textbf{DistSum$_{cosine}$}, we utilize cosine distance $\Delta(x_i, x_j) = 1 - \cos(emb(x_i), emb(x_j))$. For \textbf{DistSum$_{L2}$}, we use Euclidean distance $\Delta(x_i, x_j) = \|emb(x_i) - emb(x_j)\|^{2}_2$.

For \textbf{Partition Entropy}, we partition $\mathcal{X}^{all}$ into 1,000 clusters using K-means. Notably, unlike \citealp{han2022measuring} and the Cluster Inertia metric above, we perform clustering on the full dataset $\mathcal{X}^{all}$ rather than the currently selected subset $\mathcal{X}$.

\subsection{Details of Data Selection Strategies}
\label{app:ds}

All IT datasets in our experiments are selected from $\mathcal{X}^{all}$ and sampled over three rounds (two for Qwen) per strategy variant, unless stated otherwise. We assume these datasets have similar average sample quality, as they come from the same source without any quality filters. Additionally, the dataset size is standardized to 10,000 samples. Thus, our experiments can more accurately reflect the correlation between dataset diversity and model performance, without introducing significant confounders.

\paragraph{K-Center-Greedy} \cite{sener2017active-Kcentergreedy, chen2023maybe-Kcentergreedy, du2023mods-Kcentergreedy, wu2023self-Kcentergreedy} This strategy begins by randomly selecting a data point from the dataset $\mathcal{X}^{all}$ as the initial point of the subset $\mathcal{X}^{(s)}$. Subsequently, it iteratively computes the closest distance between the remaining points in $\mathcal{X}^{all} \setminus \mathcal{X}^{(s)}$ and selected samples in $\mathcal{X}^{(s)}$. The point with the maximum minimum distance (i.e., the farthest point) is added to $\mathcal{X}^{(s)}$. This process continues until the desired subset size is achieved.

\paragraph{Repr Filter} \cite{liu2023makes} Unlike the K-Center-Greedy strategy, which selects the farthest point from the remaining data pool, the Repr Filter randomly selects a data point whose similarity with all embeddings in $\mathcal{X}^{(s)}$ is below a predefined threshold. Due to the unique distribution of embeddings across different models, it is necessary to set distinct thresholds for each similarity function and model embedding. To ensure diversity across different experimental rounds, we employ cosine similarity and set the threshold to 0.3 for LLaMA-3-8B and 0.1 for Qwen-2.5-7B.

\paragraph{QDIT} \cite{bukharin2023data-QDIT} QDIT sampling combines diversity and quality scores for data selection; however, in our work, we focus exclusively on its diversity score. This method computes the sum of similarities between each sample in $\mathcal{X}^{all} \setminus \mathcal{X}^{(s)}$ and its closest data point in $\mathcal{X}^{(s)}$. For each candidate data point, we calculate the similarity sum as if it were added to $\mathcal{X}^{(s)}$, defining its Facility Location (FL) score. The algorithm then iteratively selects the data point with the highest FL score. For the initial selection, it chooses the data point that exhibits the highest overall similarity to all other embeddings. In our experiments, we employ cosine similarity for computing these scores. Since the Facility Location function yields a fixed subset $\mathcal{X}^{(s)}$ for a given $\mathcal{X}^{all}$, and to maintain consistency with other strategies, we utilize the same subset of data but vary the training random seeds across three rounds of experiments.

\paragraph{K-means Clustering} \cite{song2024iterselecttune-Kmeans, ge2024clustering} For this strategy, we apply the K-means clustering algorithm \cite{macqueen1967some} to partition all sample embeddings in $\mathcal{X}^{all}$ into $K$ clusters. Subsequently, given a target data budget $n$, we randomly sample $\frac{n}{K}$ data points from each cluster. For our experiments, we use both 1000 and 100 clusters for LLaMA-3-8B, and 100 clusters for Qwen-2.5-7B.

\paragraph{Random Selection} In this baseline strategy, we randomly sample 10,000 data points from $\mathcal{X}^{all}$. To explore the impact of data sources, we also sample from individual datasets, including Alpaca \cite{alpaca}, Dolly \cite{dolly}, WizardLM, UltraChat, and ShareGPT, with similar preprocessing. Although we assume that the average sample quality of these sources does not significantly differ from that of $\mathcal{X}^{all}$, we use only a single round of results from each source in the overall correlation analysis as supplementary data to avoid potential quality differences affecting the outcome.

\paragraph{Duplicate Selection} To address the challenge of defining low-diversity datasets, which is crucial for our study, we construct datasets with redundant samples. Given a target data budget $n$, the dataset is constructed by selecting $m$ unique data points, each duplicated $\frac{n}{m}$ times. We set $m$ to 1, 10, 50, 100, 500, 1000, 2000, and 5000. This approach allows us to systematically control and analyze the impact of diversity on model performance.

\subsection{Details of Performance Evaluation}
\label{app:eval}
We follow prior work in adopting LLM-based judges—MT-Bench and AlpacaEval—as they demonstrate strong alignment with human preferences and offer broad coverage of downstream tasks \cite{zhou2024lima,zhao2024long}. AlpacaEval evaluates single-turn dialogue ability through pairwise preference comparisons between model responses and those of a strong baseline, judged by GPT-4 \cite{achiam2023gpt}. MT-Bench, by contrast, assesses multi-turn conversational ability via GPT-4-based evaluations. Together, these benchmarks cover a wide range of diverse user queries and representative IT tasks, including mathematics and code generation. In our evaluation, we use GPT-4-0613 as the judge for both MT-Bench and AlpacaEval. For AlpacaEval, we follow the original setup to adopt text-davinci-003 responses as the baseline.

To account for the length bias inherent in LLM-based evaluation, we adopt the Length-Controlled Win Rates metric \cite{dubois2024length} for Alpaca-Eval, which has demonstrated stronger alignment with human judgments. We also verify that the average response lengths across the evaluated instruction-tuned (IT) models show minimal variation. This consistency is achieved by controlling for length when sampling from each model’s IT training dataset—for example, by using a fixed context length when computing embeddings and removing quality filters that might favor longer samples. Based on these measures, we believe our evaluation methodology offers a more reliable assessment of model performance.

Conversely, we didn't incorporate additional evaluation methods, such as multiple-choice QA benchmarks, because they may introduce biases toward specific domains and may not align well with human preferences. Thus, while we acknowledge the inherent limitations of LLM-based evaluation, it appears to remain the most accepted method for fairly evaluating open-ended responses in the absence of better alternatives.

\subsection{Details of Correlation Measures}
\label{app:corr}

We compute the correlation between each diversity metric and model performance using both Pearson \cite{cohen2009pearson} and Spearman \cite{zar2005spearman} correlation measures.
For example, Pearson's $r$ for a metric $\mathcal{M}_{t}$ is computed as:
\begin{equation}
    r_{\mathcal{M}_{t},\ \mathcal{P}}^{Pearson} = \frac{
        \sum_{s} (\mathcal{M}_{t}^{(s)} - \bar{\mathcal{M}}_t) (\mathcal{P}^{(s)} - \bar{\mathcal{P}})
    }{
        \sigma_{\mathcal{M}_t} \sigma_{\mathcal{P}}
    }
\end{equation}

\subsection{Details of Model Fine-Tuning}
\label{app:ft}
In our experiments, we leverage four or eight NVIDIA H800 GPUs for training the LLaMA-3-8B and Qwen-2.5-7B models. To enable efficient parallel training, we implement DeepSpeed Zero-Stage 2. Across all experiments conducted in this study, the training parameters are configured as follows: a maximum input length of 4096 tokens, a batch size of 128, 3 training epochs, a learning rate of 2e-5, and a warm-up ratio of 0.1 utilizing cosine warm-up. We use the official chat templates of LLaMA-3 and Qwen-2.5, respectively, to fine-tune each model. All models are trained with BF16 precision to optimize computational efficiency and memory usage. A single run of fine-tuning on a 10k dataset typically takes about one hour.

\section{Implementation Details}
\label{app:impl}

\subsection{Implementation Details of \textit{NovelSum}}
\label{app:impl_novelsum}
As described in Section \ref{sec:PM}, our approach to computing data diversity incorporates both proximity-weighted and density-aware considerations. In practice, we begin by embedding the samples in the given dataset \(\mathcal{X}^{(s)}\) as vectors and computing a similarity matrix that captures pairwise distances. We then apply proximity and density weights to achieve the desired outcomes.  

To estimate sample density, we utilize \textbf{FAISS} \cite{johnson2019billion}, which efficiently leverages GPU capabilities for vector similarity searches. Specifically, for each sample in \(\mathcal{X}^{(s)}\), we identify its \(k=10\) nearest neighbors within the overall sample space \(\mathcal{X}^{all}\) to compute its density factor, which we then broadcast to match the dimensions of the similarity matrix. Next, we perform element-wise multiplication between the density matrix and the similarity matrix to obtain density-aware distances in the embedding space.

Subsequently, we sort each row of the resulting matrix to determine the proximity ranks of all samples relative to the corresponding sample in that row. Finally, we compute the proximity-weighted sum for each row to derive each sample's "novelty" score and sum these scores to obtain $\mathcal{M}_{NovelSum}(\mathcal{X})$.

Notably, the density factor used in the density-aware distance computation (Eq.~\ref{eq:dad}) is $\sigma(x_j)$—that of the target point—rather than the current point’s own $\sigma(x_i)$. This choice reflects our view that applying the density factor can alter the ranking of nearest neighbors for the current point $x_i$, i.e., it changes $\pi_i(j)$, which in turn affects the outcome of the proximity-weighted sum. Using the target point’s density $\sigma(x_j)$ better captures this shift: for instance, distances to points in high-density regions are amplified, making them less likely to be considered nearest neighbors compared to those in low-density areas. In contrast, using $\sigma(x_i)$ does not alter $\pi_i(j)$ and thus fails to reflect this effect.

As noted earlier, we set the hyperparameters to $\alpha = 1$, $\beta = 0.5$, and $K = 10$ for experiments with both LLaMA-3-8B and Qwen-2.5-7B.

\subsection{Implementation Details of \textit{NovelSelect}}
\label{app:impl_novelselect}
Since selecting a subset from $\mathcal{X}^{all}$ that maximizes \textit{NovelSum} is an NP-Hard problem, similar to selecting a subset with maximum Euclidean distance, we implement a greedy strategy (Section~\ref{sec:DSS}). 

In our implementation, we iteratively compute the sample-level "novelty" $v(x)$ (Eq.~\ref{eq:v}) for each unselected candidate point with respect to the currently selected set, following the same computation process as \textit{NovelSum}. At each step, the candidate with the highest $v(x)$ is added to the subset. Notably, the density factor used for distance computation is $\sigma(x_j)$—that of the selected point—rather than the candidate’s own $\sigma(x)$, to remain consistent with \textit{NovelSum}'s definition. That said, as an alternative greedy strategy, replacing $\sigma(x_j)$ with $\sigma(x_j) + \sigma(x)$ to jointly account for both samples’ densities may also be reasonable, and we leave this for future exploration.

\section{Computational Complexity}
\label{app:complexity}

In practice, both \textit{NovelSum} and \textit{NovelSelect} incur acceptable computational costs—approximately 10 seconds and under one hour, respectively—relative to the overall fine-tuning process, and are comparable to or more efficient than many existing methods. Crucially, our approaches avoid pairwise computations over the entire large-scale source dataset of size $N$, operating instead on the selected subset, which is typically small (e.g., $n = 10{,}000$ in our experiments). And for density estimation, which considers the distribution of source samples, we leverage modern vector search libraries such as FAISS (Appendix~\ref{app:impl}). FAISS supports near-constant-time nearest neighbor queries independent of $N$, with only a one-time $O(N)$ cost to index all source samples—both negligible in the overall computation.

\textit{NovelSum} has a time complexity of $O(n^2)$ as it computes pairwise distances among the 
 selected samples. This is as efficient as most existing embedding-based diversity measures. For density estimation that accounts for source sample distribution, we use FAISS, which incurs an approximate cost of $O(N)$ for indexing all source samples and $O(n)$ for querying the selected samples' density factor—both negligible in the overall computation. In practice, computing \textit{NovelSum} (including density precomputation) on 10k samples with 4096-dimensional embeddings takes only ~10 seconds on a single H800 GPU. Additionally, the one-time cost of building the FAISS index for 396k source samples is also under 10 seconds.

For our data selection strategy, \textit{NovelSelect} runs in $O(N\cdot n^2)$ time—significantly more efficient than QDIT's $O(N^3)$. This involves distance computation between candidate samples (of size $N$) and selected samples (of size $n$) across $n$ selection iterations. In practice, selecting 10k samples from a 396k-sample pool takes under one hour using a single H800 GPU, which is faster than fine-tuning on 10k samples and negligible compared to fine-tuning on the full 396k dataset.

Embedding extraction, a shared step across embedding-based methods, takes under two hours for 396k samples in $\mathcal{X}^{all}$ using LLaMA-3-8B and vLLM on 8×H800 GPUs. As a one-time cost, this remains acceptable.

\section{Data Statistics}
\input{latex/table/data_staus}
Our data sources are detailed in Table \ref{tab:data_stastics}. After filtering out short data and non-English data, approximately 396K samples remain in $\mathcal{X}^{all}$ for use in our experiments. Note that we use the latest versions of these datasets, which may have a larger size than the initial versions. These datasets encompass samples from a wide range of domains.

\section{More Results and Analysis}

\subsection{Hyperparameter Analysis}
\label{app:hyper}
We conduct a more fine-grained hyperparameter analysis to study the effects of varying $\alpha$ and $\beta$, and investigate the sensitivity to hyperparameters for potential broader application of \textit{NovelSum}. The results are shown in Table~\ref{tbl:alpha_beta}:

\begin{table}[ht!]
\centering
\resizebox{\linewidth}{!}{
\begin{tabular}{lcc}
\toprule
\textbf{Variants} & \textbf{LLaMA-3-8B} & \textbf{Qwen-2.5-7B} \\
\midrule
NovelSum ($\alpha=1$, $\beta=0.5$) & \textbf{0.97} & \textbf{0.90} \\
\midrule
\hspace{0.10cm} - $\alpha=0$ & 0.55 & 0.51 \\
\hspace{0.10cm} - $\alpha=0.5$ & 0.77 & 0.64 \\
\hspace{0.10cm} - $\alpha=0.8$ & 0.91 & 0.85 \\
\hspace{0.10cm} - $\alpha=0.9$ & 0.94 & 0.88 \\
\hspace{0.10cm} - $\alpha=1.1$ & 0.95 & \textbf{0.91} \\
\hspace{0.10cm} - $\alpha=1.2$ & 0.93 & 0.89 \\
\hspace{0.10cm} - $\alpha=1.5$ & 0.86 & 0.86 \\
\hspace{0.10cm} - $\alpha=2$ & 0.81 & 0.82 \\
\midrule
\hspace{0.10cm} - $\beta=0$ & 0.91 & 0.73 \\
\hspace{0.10cm} - $\beta=0.2$ & 0.93 & 0.80 \\
\hspace{0.10cm} - $\beta=0.3$ & 0.94 & 0.83 \\
\hspace{0.10cm} - $\beta=0.4$ & 0.94 & 0.86 \\
\hspace{0.10cm} - $\beta=0.6$ & 0.94 & \textbf{0.91} \\
\hspace{0.10cm} - $\beta=0.7$ & 0.92 & 0.82 \\
\hspace{0.10cm} - $\beta=0.8$ & 0.88 & 0.69 \\
\hspace{0.10cm} - $\beta=1$ & 0.76 & 0.37 \\
\bottomrule
\end{tabular}
}
\caption{Hyperparameter analysis of \textit{NovelSum} with varying $\alpha$ and $\beta$ configurations on LLaMA-3-8B and Qwen-2.5-7B.}
\label{tbl:alpha_beta}
\vspace{-2mm}
\end{table}

These results show that NovelSum consistently achieves strong correlation with model performance across a relatively wide range of hyperparameters, without drastic fluctuations. This suggests that the sensitivity issue may not be particularly severe in practice.

From a theoretical perspective, we view the proximity decay coefficient $\alpha$ as related to the semantic richness of the source data. For richer dataset, it's better to consider more neighbors in distance computations, corresponding to a smaller $\alpha$. Given that most current IT tasks and datasets are semantically rich, the current choice of $\alpha$ is likely to remain effective as long as the domain is not overly narrow. On the other hand, the density coefficient $\beta$ controls the balance between distance and density components. We believe this balance is not specific to a particular dataset, but rather general across IT tasks. While the exact optimal value of $\beta$ may vary slightly depending on the implementation of distance and density calculations, the use of cosine distance and nearest-neighbor density estimation—as adopted in our work—provides a stable basis. Therefore, re-tuning $\beta$ is unlikely to be necessary in most cases.

Based on the above discussion, we believe our current hyperparameter configuration is robust for general instruction tuning and can exhibit a considerable degree of generalizability across broader scenarios. This helps reduce the need for costly hyperparameter tuning. Therefore, the level of hyperparameter sensitivity observed here may not be a major obstacle to the broader applicability of our method.

\subsection{Analysis of Embedding Extractors}
\label{app:embed}
To investigate the impact of the embedding model choice on diversity measurements, we conduct an additional ablation study using four different models to generate embeddings for \textit{NovelSum} computation: LLaMA-3-8B, LLaMA-2-7B, Qwen-2.5-7B, and BERT-base. We then measure the correlation between the resulting \textit{NovelSum} scores and instruction tuning (model) performance under two fine-tuning backbones: LLaMA-3-8B and Qwen-2.5-7B.

\begin{table}[ht!]
\centering
\resizebox{\linewidth}{!}{
\begin{tabular}{lcc}
\toprule
\multirow{2}{*}{\textbf{Embedding Model}} & \multicolumn{2}{c}{\textbf{Fine-tuning Backbone}} \\
\cmidrule(lr){2-3}
& \textbf{LLaMA-3-8B} & \textbf{Qwen-2.5-7B} \\
\midrule
LLaMA-3-8B & \textbf{0.97} & 0.81 \\
Qwen-2.5-7B & 0.92 & \textbf{0.90} \\
LLaMA-2-7B & 0.94 & 0.87 \\
BERT-base & 0.90 & 0.64 \\
\bottomrule
\end{tabular}
}
\caption{Correlation between \textit{NovelSum} computed using different embedding models and instruction tuning performance under two fine-tuning backbones.}
\label{tbl:embed_ablation}
\vspace{-2mm}
\end{table}

The results, presented in Table~\ref{tbl:embed_ablation}, indicate that using the same model for both embedding extraction and fine-tuning yields the strongest correlation between diversity metrics and instruction tuning performance, likely due to shared representation spaces. In contrast, employing a general-purpose encoder such as BERT-base leads to weaker correlations compared to other LLMs.

Following prior work, we use pretrained base models directly (Appendix~\ref{app:preprocess}) for sample embedding computation, primarily for research purposes. For practical applications, however, one may consider using state-of-the-art LLM-based embedding models fine-tuned specifically for embedding tasks, which may offer improved performance.

\subsection{Theoretical Analysis}
\label{app:theo}

We begin by situating our work within broader literatures. The problem of selecting representative objects from a given set has been extensively studied in Operations Research \cite{ravi1994heuristic, fekete2004maximum, cevallos2017local}, often through formulations such as maximum dispersion and facility location. These approaches share similar motivations with our method and help explain the effectiveness of the density-aware distance.
In parallel, prior work on sampling strategies \cite{eldar1997farthest} conceptualizes sampling as a stochastic process for reconstruction and highlights the effectiveness of maximizing inter-sample distances in progressive image sampling.
Building on similar insights, our diversity metric \textit{NovelSum} accounts for both inter-sample distances and information density in the sample space.

To facilitate a deeper understanding of \textit{NovelSum}, we offer the following theoretical interpretation as a possible perspective: The basic sum of semantic distances among all samples (Eq.~\ref{eq:distsum}) represents the semantic variance of the selected samples. A larger variance implies significant differences among some samples but does not necessarily guarantee a diverse semantic distribution across the entire semantic space. In contrast, the \textbf{proximity-weighted sum} (Eq.~\ref{eq:pws}) specifically measures semantic diversity by focusing on the gaps between neighboring samples rather than on distant ones. A larger proximity-weighted sum indicates that samples are more distinct from their immediate neighbors, reflecting higher overall semantic diversity. Further, the \textbf{density-aware distance} (Eq.~\ref{eq:dad}) incorporates an additional density factor into the distance calculation, explicitly considering information density within the semantic space. By extracting information from the gaps between neighboring samples—multiplying the semantic distance between each sample and its neighbors by information density factors and performing a proximity-weighted sum—we effectively quantify each sample’s unique informational contribution. Consequently, \textit{NovelSum} measures the total unique information of all samples, given their semantic embeddings and scenario-specific information density (e.g., general instruction tuning). This aggregate value corresponds to the "IT-aligned Diversity" we aim to measure.

\subsection{Additional Results}
\label{app:results}
\begin{figure}[b]
    \centering
        \includegraphics[width=\linewidth]{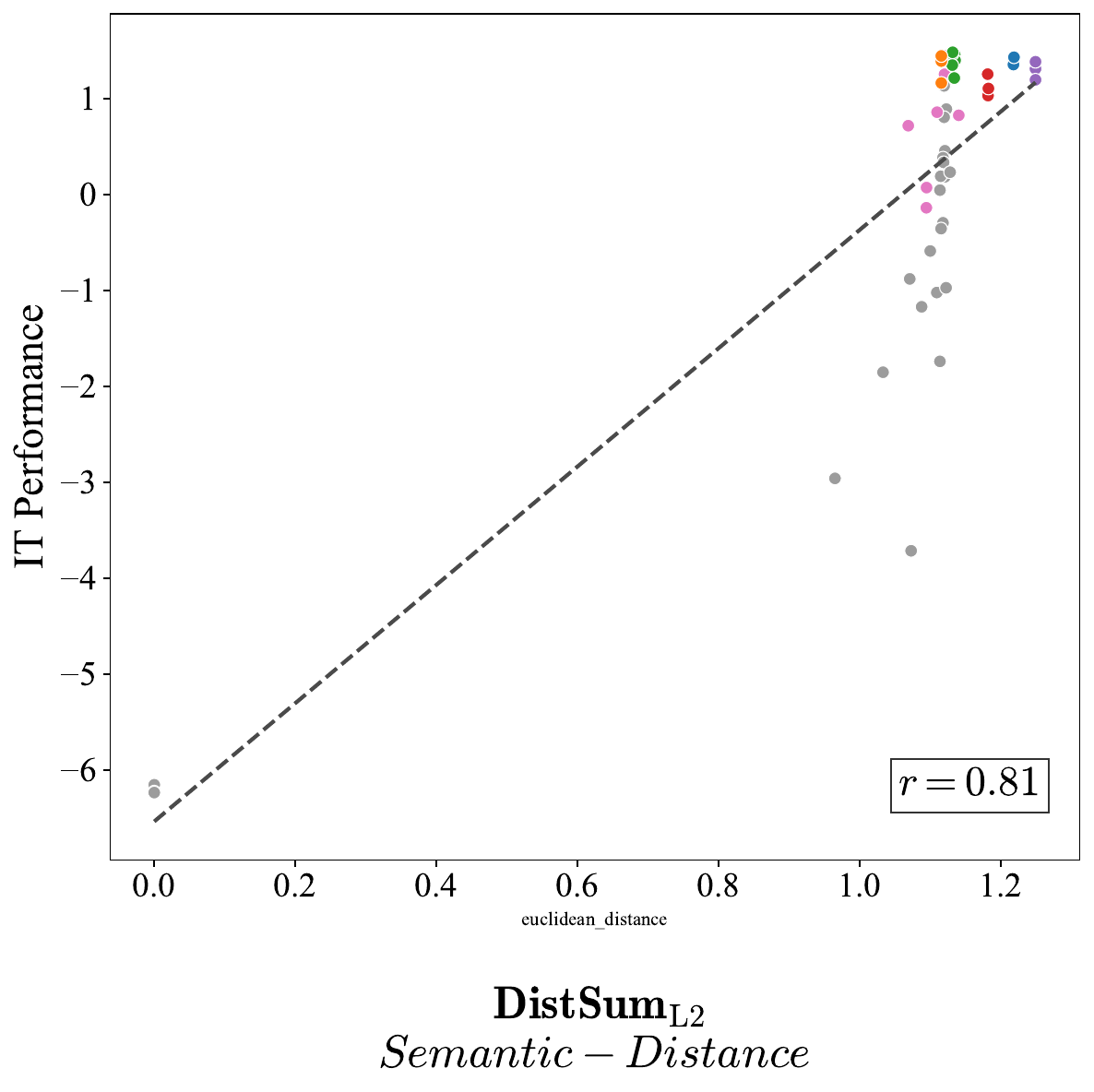}
    \caption{Evaluation of DistSum$_{L2}$ metric by their correlation with IT performance.}
    \label{fig:scatter_l2_appdenix}
\end{figure}

\begin{figure}[b]
    \centering
        \includegraphics[width=\linewidth]{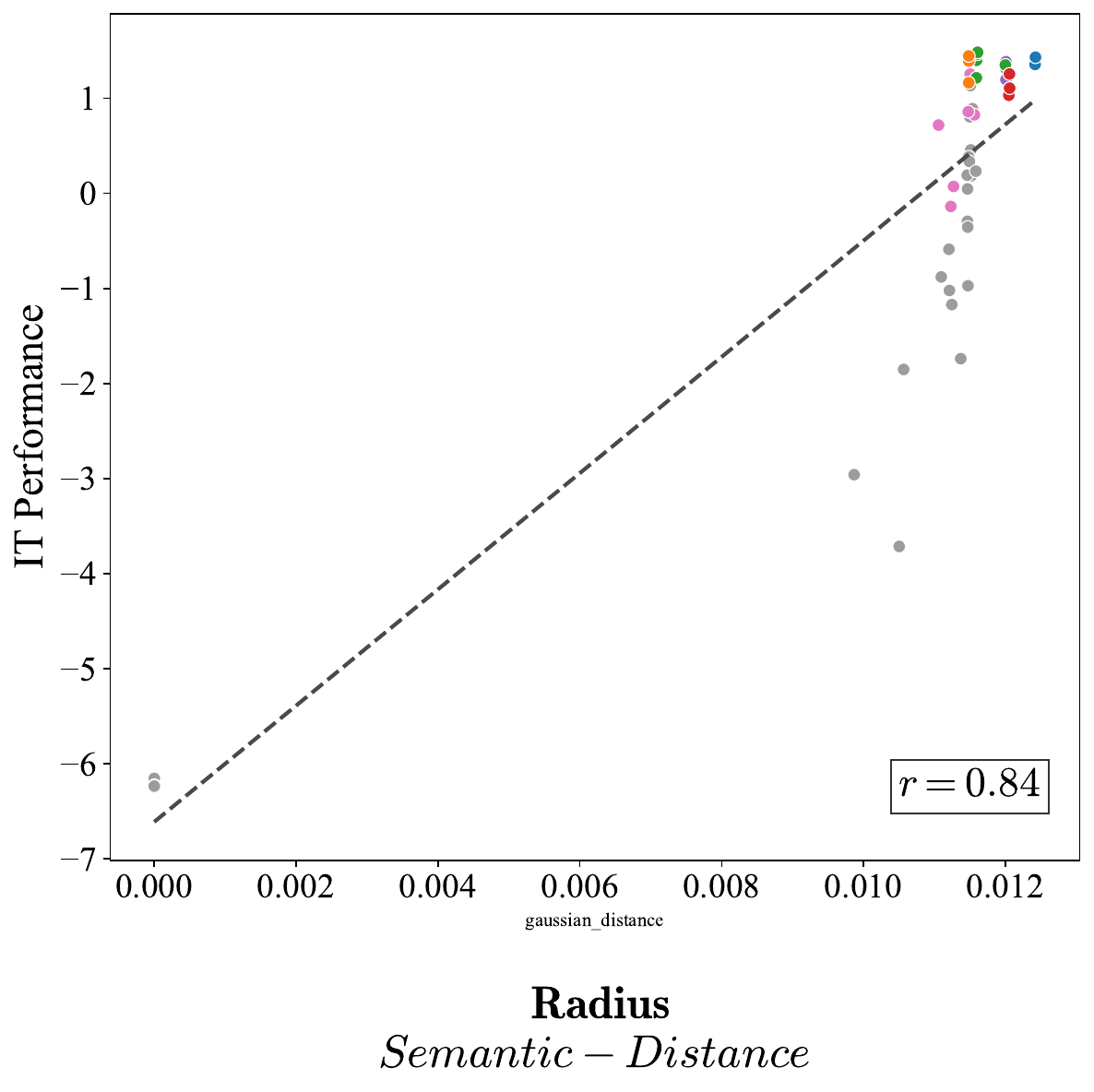}
    \caption{Evaluation of Radius metric by their correlation with IT performance.}
    \label{fig:scatter_radius_appdenix}
\end{figure}

\begin{figure}[b]
    \centering
        \includegraphics[width=\linewidth]{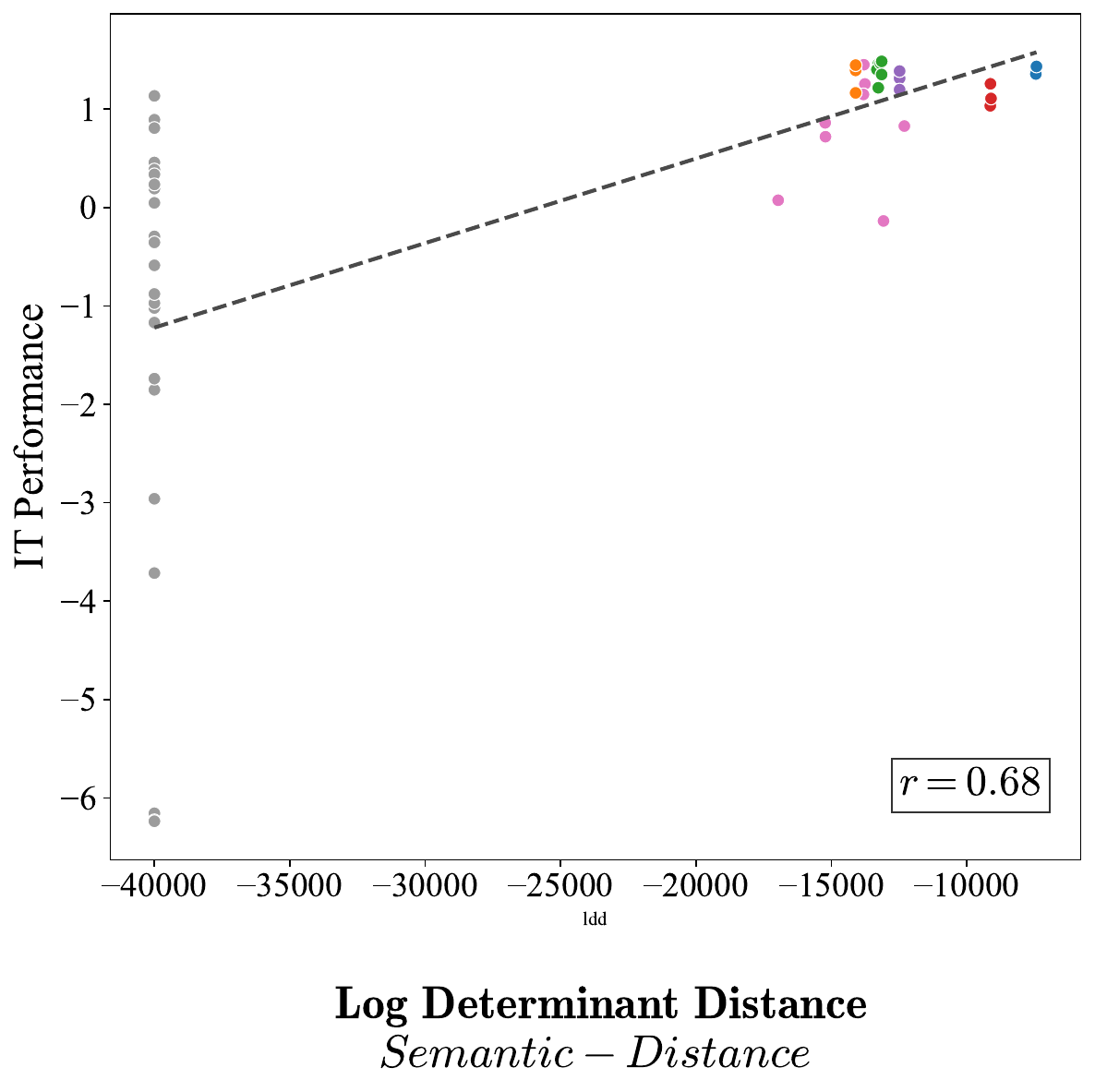}
    \caption{Evaluation of Log Determinant Distance metric by their correlation with IT performance.}
    \label{fig:scatter_ldd_appdenix}
\end{figure}

Additional scatter plots for the analysis in Section \ref{sec:existing} are provided in Figure \ref{fig:scatter_l2_appdenix}, Figure \ref{fig:scatter_radius_appdenix} and Figure \ref{fig:scatter_ldd_appdenix} , illustrating the correlation for DistSum$_{L2}$, Radius, and Log Determinant Distance, respectively.

The full results of the correlation experiments on LLaMA-3-8B and Qwen-2.5-7B are presented in Table \ref{tab:more_results_llama3} and Table \ref{tab:more_results_qwen}, respectively. These tables provide a comprehensive comparison of diversity metrics across different experimental configurations.

\section{Others}
\subsection{License for Artifacts and Data Consent}
In this paper, the artifacts used are all available for academic research work, including ShareGPT, WizardLM, UltraChat, Alpaca and Dolly.
The diversity metrics and data selection methods compared in this paper can all be used for academic research.
All data originates from the original authors' open-source releases and can be used for academic research and publication.

\subsection{Data Statement}

The training datasets may contain offensive content; however, they do not include any personal information.
Furthermore, our approach is designed to drive the instruction tuning of the model toward better alignment with human preferences, thereby mitigating the generation of harmful content.

\subsection{AI Assistant Usage Statement}

We utilized ChatGPT for writing refinement and minor coding assistance. AI assistants were not employed for research innovation, and all core contributions were solely developed by the authors.

\input{latex/table/appendix_llama_table}
\input{latex/table/appendix_qwen_table}

\subsection{Budgets}
\label{ap:budgets}
We instruction-tune (train) each model for approximately one hour on a single node with eight H800-80G GPUs, totaling around 80 hours across 80 runs. Additionally, we spend around \$1,000 on the GPT API to evaluate our models using MT-bench and AlpacaEval.


%% file: latex/table/data_staus.tex
\begin{table}[t!]
\centering
\resizebox{0.88\columnwidth}{!}{
\begin{tabular}{lcr}
\toprule
\textbf{Data Pool}           & \textbf{Dataset Source}   & \textbf{Sample Size} \\ 
\midrule
\multirow{3}{*}{$\mathcal{X}^{all}$} & ShareGPT                  & 103 K                 \\
                             & UltraChat                 & 207 K                \\
                             & WizardLM & 196 K                \\
 \midrule
\multirow{2}{*}{$\mathcal{X}^{other}$} & Alpaca                    & 52 K                 \\
                             & Dolly                     & 15 K                 \\
\bottomrule
\end{tabular}}
\caption{Statistics of Data Pools $\mathcal{X}^{all}$ and $\mathcal{X}^{other}$. 
The column "Dataset Source" indicates the origin of the data used for sampling, while "Sample Size" denotes the number of samples in each dataset. This table provides an overview of the data used in our experiments.}
\label{tab:data_stastics}
\vspace{-2mm}
\end{table}

%% file: latex/table/appendix_llama_table.tex
\begin{table*}[t!]
  \centering
  \resizebox{\linewidth}{!}{
  \begin{tabular}{lc|ccc|cccccccccccc}
    \toprule
    \textbf{Data Selection Strategy} & \textbf{Runs} & \textbf{Alpaca} & \textbf{MT - bench} & \textbf{Aggregated} & \textbf{NovelSum} & \textbf{DistSum$_{cosine}$} & \textbf{DistSum$_{L2}$} & \textbf{KNN} & \textbf{Inertia} & \textbf{Radius} & \textbf{VS} & \textbf{Entropy} & \textbf{FL} & \textbf{LDD} & \textbf{TTR} & \textbf{vocd - D} \\
    &  &  &  &  &  &  &  &  &  & \textbf{$10^{-2}$} & \textbf{$10^7$} &  & \textbf{$10^5$} & \textbf{$10^4$} &  &  \\
    \midrule
    Random - alpaca & 1 & 58.8 & 4.40 & 0.07 & 0.572 & 0.605 & 1.09 & 0.509 & 0.293 & 1.13 & 1.32 & 8.93 & 2.83 & -1.70 & 0.853 & 82.3 \\
    Random - dolly   & 1 & 47.5 & 4.34 & -0.14 & 0.500 & 0.603 & 1.09 & 0.596 & 0.362 & 1.12 & 1.70 & 7.83 & 2.59 & -1.31 & 0.844 & 76.6 \\
    Random - sharegpt & 1 & 74.6 & 6.46 & 0.83 & 0.628 & 0.656 & 1.14 & 0.574 & 0.380 & 1.16 & 1.70 & 8.97 & 2.83 & -1.23 & 0.850 & 78.3 \\
    Random - ultrachat  & 1 & 72.5 & 6.76 & 0.86 & 0.672 & 0.622 & 1.11 & 0.567 & 0.323 & 1.15 & 1.48 & 9.40 & 2.96 & -1.52 & 0.881 & 110 \\
    Random - wizardlm  & 1 & 76.9 & 5.82 & 0.72 & 0.591 & 0.578 & 1.07 & 0.594 & 0.331 & 1.11 & 1.44 & 9.08 & 2.88 & -1.52 & 0.858 & 85.7 \\
    Random - $\mathcal{X}^{all}$  & 3 & 75.5 & 6.18 & 1.20 & 0.675 & 0.634 & 1.12 & 0.606 & 0.353 & 1.15 & 1.61 & 9.80 & 2.99 & -1.38 & 0.870 & 97.2 \\
    Farthest  & 3 & 74.0 & 6.30 & 1.22 & 0.687 & 0.789 & 1.25 & 0.407 & 0.350 & 1.20 & 1.56 & 6.52 & 2.14 & -1.25 & 0.837 & 68.3 \\
    Duplicate m = 1 & 3 & 0.57 & 1.01 & -6.36 & 0.000 & 0.000 & 0.00 & 0.000 & 0.000 & 0.00 & 1.08 & 0.00 & 1.25 & -inf & 0.887 & 121 \\
    Duplicate m = 10  & 3 & 31.2 & 3.54 & -2.97 & 0.268 & 0.589 & 1.02 & 0.000 & 0.000 & 1.03 & 7.16 & 3.27 & 2.08 & -inf & 0.863 & 90.0 \\
    Duplicate m = 50   & 3 & 51.0 & 4.38 & -1.35 & 0.388 & 0.608 & 1.09 & 0.001 & 0.000 & 1.12 & 2.74 & 5.58 & 2.40 & -inf & 0.873 & 101 \\
    Duplicate m = 100  & 3 & 63.6 & 5.42 & 0.05 & 0.461 & 0.634 & 1.12 & 0.001 & 0.000 & 1.15 & 4.95 & 6.50 & 2.52 & -inf & 0.866 & 92.3 \\
    Duplicate m = 500   & 3 & 65.0 & 5.67 & 0.30 & 0.556 & 0.635 & 1.12 & 0.001 & 0.222 & 1.15 & 1.79 & 8.47 & 2.75 & -inf & 0.869 & 96.7 \\
    Duplicate m = 1000  & 3 & 71.3 & 6.00 & 0.86 & 0.587 & 0.630 & 1.12 & 0.001 & 0.292 & 1.15 & 2.99 & 9.06 & 2.83 & -inf & 0.869 & 96.1 \\
    Duplicate m = 2000 & 3 & 59.6 & 5.31 & -0.23 & 0.618 & 0.633 & 1.12 & 0.001 & 0.330 & 1.15 & 5.07 & 9.46 & 2.90 & -inf & 0.870 & 97.3 \\
    Duplicate m = 5000  & 3 & 51.5 & 4.85 & -0.98 & 0.656 & 0.634 & 1.12 & 0.001 & 0.349 & 1.15 & 9.92 & 9.72 & 2.97 & -inf & 0.871 & 97.1 \\
    K - Center - Greedy  & 3 & 75.3 & 6.33 & 1.31 & 0.687 & 0.746 & 1.22 & 0.864 & 0.522 & 1.24 & 2.53 & 9.30 & 2.73 & -7.44 & 0.862 & 88.5 \\
    Kmeans Clustering$_{1000}$  & 3 & 76.5 & 6.31 & 1.35 & 0.692 & 0.646 & 1.13 & 0.615 & 0.372 & 1.17 & 1.70 & 9.87 & 2.99 & -1.32 & 0.869 & 96.4 \\
    Kmeans Cluster$_{100}$   & 3 & 74.4 & 6.36 & 1.28 & 0.693 & 0.650 & 1.13 & 0.610 & 0.362 & 1.16 & 1.69 & 9.78 & 2.99 & -1.33 & 0.869 & 96.1 \\
    QDIT   & 3 & 75.9 & 6.21 & 1.25 & 0.673 & 0.629 & 1.12 & 0.602 & 0.348 & 1.15 & 1.59 & 9.77 & 2.99 & -1.41 & 0.871 & 98.5 \\
    Repr \ Filter & 3 & 72.6 & 6.17 & 1.05 & 0.671 & 0.703 & 1.18 & 0.799 & 0.470 & 1.21 & 2.23 & 9.45 & 2.86 & -9.12 & 0.866 & 92.0 \\
    NoveSelect  & 3 & 78.1 & 6.47 & 1.55 & 0.762 & 0.821 & 1.28 & 0.704 & 0.534 & 1.30 & 2.55 & 9.23 & 2.73 & -6.27 & 0.862 & 87.9 \\
    \bottomrule
  \end{tabular}
  }
  \caption{Comprehensive experimental results on LLaMA-3-8B. Each data selection strategy variant is evaluated over three independent runs (except for random selection) to ensure the robustness and reliability of the findings. The results from multiple runs are averaged. Note that \textit{NovelSelect} results are only included in Section \ref{sec:DSS} and are not part of the correlation calculations. Details of the data selection strategies are provided in Appendix \ref{app:ds}.}
  \label{tab:more_results_llama3}
\end{table*}

%% file: latex/table/appendix_qwen_table.tex
\begin{table*}[t!]
  \centering
  \resizebox{\linewidth}{!}{
  \begin{tabular}{lc|ccc|cccccccccccc}
    \toprule
    \textbf{Data Selection Strategy} & \textbf{Runs} & \textbf{Alpaca} & \textbf{MT-bench} & \textbf{Aggregated} & \textbf{NovelSum} & \textbf{DistSum$_{cosine}$} & \textbf{DistSum$_{L2}$} & \textbf{KNN} & \textbf{Inertia} & \textbf{Radius} & \textbf{VS} & \textbf{Entropy} & \textbf{FL} & \textbf{LDD} & \textbf{TTR} & \textbf{vocd-D} \\
    &  &  &  &  &  &  &  &  &  & \textbf{$10^{-3}$} & \textbf{$10^7$} &  & \textbf{$10^5$} & \textbf{$10^4$} &  &  \\
    \midrule
    Random - alpaca & 1 & 71.5 & 5.52 & -0.38 & 0.336 & 0.189 & 0.596 & 0.223 & 0.066 & 4.06 & 1.89 & 8.66 & 3.50 & -4.40 & 0.853 & 82.4 \\
    Random - dolly & 1 & 55.2 & 6.24 & -0.49 & 0.320 & 0.221 & 0.651 & 0.293 & 0.098 & 4.52 & 3.04 & 7.92 & 3.46 & -3.62 & 0.844 & 76.6 \\
    Random - sharegpt & 1 & 82.4 & 7.91 & 0.07 & 0.392 & 0.285 & 0.731 & 0.289 & 0.110 & 4.64 & 3.36 & 8.87 & 3.51 & -3.34 & 0.850 & 78.3 \\
    Random - ultrachat & 1 & 78.0 & 7.66 & -0.02 & 0.389 & 0.200 & 0.620 & 0.252 & 0.074 & 4.11 & 2.09 & 9.30 & 3.52 & -4.17 & 0.881 & 110 \\
    Random - wizardlm & 1 & 77.1 & 7.28 & -0.08 & 0.349 & 0.211 & 0.631 & 0.296 & 0.093 & 4.42 & 2.65 & 9.03 & 3.50 & -3.84 & 0.858 & 85.7 \\
    Random $\mathcal{X}^{all}$ & 2 & 81.9 & 7.57 & 0.87 & 0.408 & 0.230 & 0.661 & 0.286 & 0.092 & 4.36 & 1.41 & 9.77 & 3.54 & -3.81 & 0.869 & 97.0 \\
    Duplicate m=50 & 2 & 69.3 & 7.41 & -1.46 & 0.252 & 0.215 & 0.638 & 0.001 & 0.000 & 4.29 & 0.13 & 5.64 & 3.44 & -inf & 0.871 & 98.1 \\
    Duplicate m=100 & 2 & 75.1 & 7.46 & -0.43 & 0.309 & 0.243 & 0.664 & 0.001 & 0.000 & 4.32 & 0.20 & 6.54 & 3.48 & -inf & 0.870 & 97.7 \\
    Duplicate m=500 & 2 & 72.9 & 7.51 & -0.69 & 0.357 & 0.240 & 0.672 & 0.001 & 0.057 & 4.41 & 0.51 & 8.50 & 3.54 & -inf & 0.868 & 94.9 \\
    Duplicate m=1000 & 2 & 78.6 & 7.59 & 0.37 & 0.364 & 0.229 & 0.658 & 0.001 & 0.076 & 4.36 & 0.74 & 9.05 & 3.56 & -inf & 0.869 & 97.0 \\
    Duplicate m=5000 & 2 & 82.0 & 7.53 & 0.81 & 0.399 & 0.230 & 0.661 & 0.001 & 0.091 & 4.36 & 1.27 & 9.68 & 3.57 & -inf & 0.870 & 97.8 \\
    K-Center-Greedy & 2 & 81.6 & 7.90 & 1.45 & 0.505 & 0.440 & 0.923 & 0.501 & 0.214 & 6.13 & 3.09 & 8.50 & 3.42 & -2.29 & 0.837 & 68.6 \\
    K-means Clustering & 2 & 79.8 & 7.84 & 1.06 & 0.440 & 0.260 & 0.698 & 0.301 & 0.106 & 4.54 & 1.60 & 9.86 & 3.54 & -3.63 & 0.868 & 94.9 \\
    QDIT & 2 & 80.0 & 7.81 & 1.00 & 0.403 & 0.223 & 0.650 & 0.283 & 0.091 & 4.33 & 2.60 & 9.74 & 3.54 & -3.87 & 0.871 & 99.1 \\
    Repr \ Filter & 2 & 81.8 & 7.83 & 1.35 & 0.495 & 0.421 & 0.901 & 0.476 & 0.199 & 5.94 & 7.15 & 8.59 & 3.46 & -2.42 & 0.839 & 69.8 \\
    \bottomrule
  \end{tabular}
  }
  \caption{Comprehensive experimental results on Qwen-2.5-7B. Each data selection strategy variant is evaluated over two independent runs (except for random selection).}
  \label{tab:more_results_qwen}
\end{table*}